\documentclass[10pt,twocolumn,letterpaper]{article}

\usepackage{3dv}
\makeatletter
\@namedef{ver@everyshi.sty}{}
\makeatother
\usepackage[nointegrals]{wasysym}
\usepackage{booktabs}
\usepackage{balance}
\usepackage{multirow}
\usepackage{overpic}
\usepackage{bbm}
\usepackage{dsfont}
\usepackage[font=small,labelfont=bf]{caption}
\usepackage{color}
\usepackage{pifont}
\usepackage{etoolbox}
\usepackage{float}
\usepackage{tikz}
\usepackage{times}
\usepackage{epsfig}
\usepackage{graphicx}
\usepackage{amsmath}
\usepackage{amssymb}
\usepackage{comment}
\usepackage{soul}
\usepackage{listings}
\usepackage{lscape}
\usepackage{rotating}
\usepackage{enumitem}
\usepackage[utf8]{inputenc}
\usepackage{pgfplots}
\DeclareUnicodeCharacter{2212}{−}
\usepgfplotslibrary{groupplots,dateplot}
\usetikzlibrary{patterns,shapes.arrows}
\pgfplotsset{compat=newest}

\definecolor{palette_sinopia}{RGB}{0,84,159}
\definecolor{palette_green}{RGB}{132,177,88}
\definecolor{my_orange}{HTML}{DEC402}
\definecolor{my_blue}{HTML}{1F77B4}

\newcommand{\refsec}[1]{Sec.\,\ref{sec:#1}}

\newcommand{\reffig}[1]{Fig.\,\ref{fig:#1}}
\newcommand{\reftab}[1]{Tab.\,\ref{tab:#1}}

\lstdefinestyle{mypython}{
  language=python,
  breaklines=true,
  basicstyle=\fontsize{8}{12}\selectfont\ttfamily,
  keywordstyle=\bfseries\color{my_blue},
  linewidth=.99\textwidth,
}

\definecolor{m_green}{RGB}{212, 251, 121}
\definecolor{m_orange}{RGB}{255, 212, 121}
\definecolor{m_red}{RGB}{255, 126, 121}
\definecolor{m_violet}{RGB}{215, 131, 255}
\definecolor{m_blue}{RGB}{118, 214, 255}
\newcommand{\colorsquare}[1]{{\color{#1}$\blacksquare$}\hspace{-7.78pt}$\square$}

\newcommand{\circlenum}[1]{{\textcircled{\footnotesize{#1}}}}

\newcommand{\parag}[1]{\vskip8pt \noindent \textbf{#1}}

\newcommand{\namelong}{Mix3D}
\newcommand{\name}{Mix3D}
\newcommand{\papertitle}{Mix3D: Out-of-Context Data Augmentation for 3D Scenes}

\definecolor{darkgreen}{RGB}{0,255,0}
\definecolor{linkgreen}{RGB}{52,130,48}

\newcommand{\ColorMapCircle}{\ding{108}}

\newcommand{\cmark}{\ding{51}}%
\newcommand{\xmark}{\ding{55}}%

\newcommand\ArrowDown[1]{
\hspace{-12px}\rotatebox[origin=c]{270}{$\curvearrowright$}{\hspace{2px}#1}
} 

\usepackage{times}
\usepackage{epsfig}
\usepackage{graphicx}
\usepackage{amsmath}
\usepackage{amssymb}
\usepackage{subcaption}
\usepackage{array,multirow}
\usepackage{xfrac}
\usepackage{color, colortbl}
\usepackage{tcolorbox}

\definecolor{LightCyan}{rgb}{0.87,0.92,0.96}
\definecolor{m_green}{RGB}{212, 251, 121}
\definecolor{m_orange}{RGB}{255, 212, 121}
\definecolor{m_red}{RGB}{255, 126, 121}
\definecolor{m_violet}{RGB}{215, 131, 255}
\definecolor{m_blue}{RGB}{184, 215, 241}

\usepackage[pagebackref=true,breaklinks=true,colorlinks,bookmarks=false]{hyperref}
\usepackage[bottom]{footmisc}

\threedvfinalcopy
\def\threedvPaperID{26}

\ifthreedvfinal\fi
\begin{document}

\title{\papertitle{}}

\newcommand{\footremember}[2]{%
   \thanks{#2}
    \newcounter{#1}
    \setcounter{#1}{\value{footnote}}%
}
\newcommand{\footrecall}[1]{%
    \footnotemark[\value{#1}]%
} 
\hyphenation{Minkowski-Net} 
\author{
\vspace{-20px}\\
Alexey Nekrasov$^{1*}$~~~~
Jonas Schult$^{1*}$~~~~
Or Litany$^{2}$~~~~
Bastian Leibe$^{1}$~~~~
Francis Engelmann$^{1,3}$\\
\vspace{-0.25cm}\\
$^{1}$RWTH Aachen University~~~~~
$^{2}$NVIDIA~~~~~
$^{3}$ETH AI Center
\vspace{-1.1cm}
}


\twocolumn[{
\renewcommand\twocolumn[1][]{#1}
\maketitle
\thispagestyle{empty}
\center

\vspace{0.5cm}
}]

\let\thefootnote\relax
\footnotetext{$*$ \text{Equal contribution.}}

\newcommand{\expPopulationLA}{$67.9$} %
\newcommand{\expPopulationHA}{$67.7$} %
\newcommand{\expPopulationLS}{$69.2$} %
\newcommand{\expPopulationHS}{$68.4$} %

\newcommand{\expCutoutAA}{$67.4$} %
\newcommand{\expCutoutBA}{$67.6$} %
\newcommand{\expCutoutCB}{$67.9$} %
\newcommand{\expCutoutDB}{$66.2$} %
\newcommand{\expCutoutEC}{$66.4$} %
\newcommand{\expCutoutFC}{$65.4$} %

\newcommand{\baseline}{$66.6$} %

\newcommand{\expMergedScannetMinkowski}{$69.0$ {\footnotesize $\pm$ $0.32$}} %
\newcommand{\expBaselineScannetMinkowski}{$66.6$ {\footnotesize $\pm$ $0.09$}} %
\newcommand{\expBaselineStanfordMinkowski}{$64.7$ {\footnotesize $\pm$ $0.34$}} %
\newcommand{\expMergedStanfordMinkowski}{$65.4$ {\footnotesize  $\pm$ $0.37$}} %

\newcommand{\expBaselineKittiMinkowski}{$56.7$} %
\newcommand{\expMergedKittiMinkowski}{$59.9$} %

\newcommand{\expBaselineScannetKPConvDeform}{$69.3$ {\footnotesize $\pm$ $0.10$}} %
\newcommand{\expMergedScannetKPConvDeform}{$\mathbf{70.3}$ {\footnotesize $\pm$ $0.06$}} %

\newcommand{\expBaselineScannetKPConvRigid}{$68.8$ {\footnotesize $\pm$ $0.10$}} %
\newcommand{\expMergedScannetKPConvRigid}{$\mathbf{69.5}$ {\footnotesize $\pm$ $0.33$}} %

\newcommand{\expBaselineStanfordKPConvRigid}{$65.0$ {\footnotesize $\pm$ $0.31$}} %
\newcommand{\expMergedStanfordKPConvRigid}{$\mathbf{66.5}$ {\footnotesize $\pm$ 0.44}} %

\newcommand{\expBaselineStanfordKPConvDeform}{$65.8$ {\footnotesize $\pm$ $0.47$}} %
\newcommand{\expMergedStanfordKPConvDeform}{$\mathbf{67.2}$ {\footnotesize $\pm$ $0.28$}} %

\newcommand{\expInstanceValBase}{$24.5$ {\footnotesize $\pm$ $0.88$}} %
\newcommand{\expInstanceValMerged}{$36.0$ {\footnotesize $\pm$ $0.57$}} %

\newcommand{\expNearby}{$67.9$} %

\newcommand{\expMergedThree}{$68.6$} %
\newcommand{\expMergedFour}{$68.7$} %
\newcommand{\expMergedSeven}{$68.3$} %
\newcommand{\expMergedEight}{$65.7$} %

\newcommand{\expNoise}{$51.4$} %
\newcommand{\expNoiseResempling}{$45.5$} %

\newcommand{\expNoiseStructured}{$68.4$} %

\newcommand{\expBigCropsBase}{$66.9$} %
\newcommand{\expBigCropsMerged}{$67.4$} %

\newcommand{\expSizedCropsBase}{$66.9$} %
\newcommand{\expSizedCropsMerged}{$67.4$} %


\begin{abstract}
We present \name{}, a data augmentation technique for segmenting large-scale 3D scenes. 
Since scene context helps reasoning about object semantics, current works focus on models with large capacity and receptive fields that can fully capture the global context of an input 3D scene.
However, strong contextual priors can have detrimental implications like mistaking a pedestrian crossing the street for a car.
In this work, we focus on the importance of balancing global scene context and local geometry, with the goal of generalizing beyond the contextual priors in the training set.
In particular, we propose a \emph{``mixing"} technique which creates new training samples by combining two augmented scenes.
By doing so, object instances are implicitly placed into novel out-of-context environments, therefore making it harder for models to rely on scene context alone, and instead infer semantics from local structure as well.
We perform detailed analysis to understand the importance of global context, local structures and the effect of mixing scenes.
In experiments, we show that models trained with Mix3D profit from a significant performance boost on indoor (ScanNet, S3DIS) and outdoor datasets (SemanticKITTI).
Mix3D can be trivially used with any existing method, \eg,
trained with Mix3D, MinkowskiNet outperforms all prior state-of-the-art methods by a significant margin on the ScanNet test benchmark ($78.1\%$ mIoU).
Code is available at: \url{https://nekrasov.dev/mix3d/}
\end{abstract}


\section{Introduction}
In this work, we address the task of 3D semantic segmentation which assigns a semantic label to every point in a given 3D point cloud.
Popularized by PointNet\,\cite{Qi17CVPR}, the broader field of deep learning for 3D scene understanding has experienced tremendous growth.
While these early approaches\,\cite{Qi17CVPR, Qi17NIPS} can classify individual objects or segment larger scenes by splitting them into cubical (or spherical) chunks,
recent high-capacity models such as MinkowskiNets\,\cite{Choy2019CVPR} and SparseConvNets\,\cite{Graham18CVPR} 
have large receptive fields and can directly process full rooms or even outdoor scenes without splitting them up.
By doing so, they can capture the global \emph{context} of a scene.
Context refers to the strong configuration rules present in man-made environments\,\cite{oliva07cogn}.
These rules manifest in re-occurrences of object arrangements.
For example, based on context, we can expect to see chairs around a table situated in a conference room.
Importantly, context supports reasoning about object semantics, as confirmed by research in visual cognition, computer vision, and cognitive neuroscience\,\cite{Biederman72arxiv, Hock74Perception, oliva07cogn}.

\begin{figure}[t]
     \centering
     \begin{tabular}{ccc}
     & Outdoor Scene & Indoor Scene\\
     & \hspace{3.5cm} & \hspace{3.5cm} \vspace{-4.5mm}\\
     \end{tabular}     
     \rotatebox{90}{\hspace{5mm} \footnotesize with Mix3D \hspace{1.15cm} without Mix3D}
     \includegraphics[width=0.96\columnwidth]{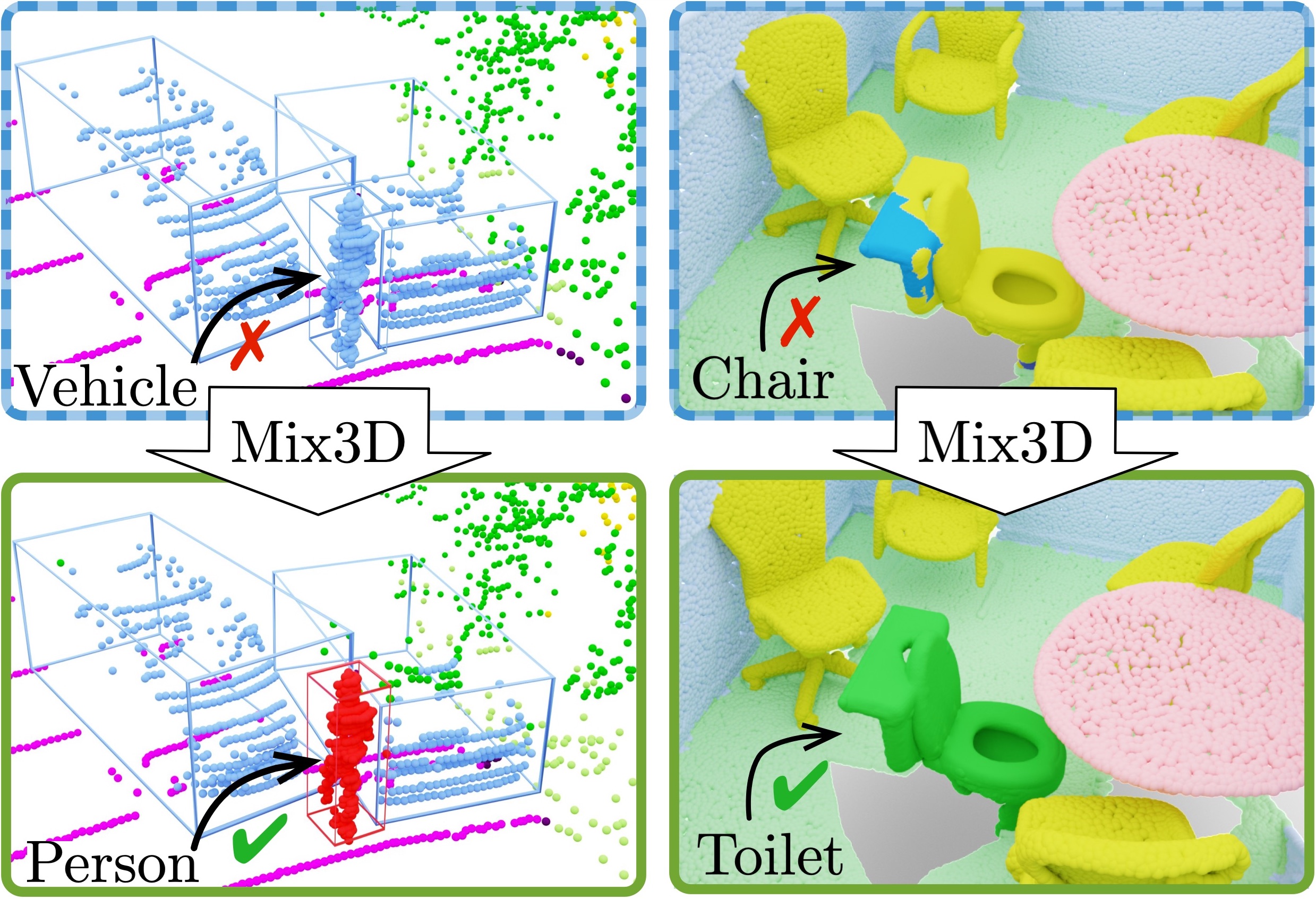}
     \caption{\small
     \textbf{Semantic segmentation improvements using Mix3D.}
     Mix3D is a data augmentation technique 
     that increases generalization beyond the contextual priors of the training scenes and improves predictions for rare events not well captured by the training data.
     Models trained with Mix3D \emph{(bottom)} can handle critical situations where pedestrians walk in the middle of the road \emph{(left)},
     and can correctly identify objects in unusual settings \emph{(right)}.}
     \label{fig:teaser}
     \vspace{-7px}
 \end{figure}

However, while contextual information has proven useful, relying too heavily on global scene priors can be harmful when reasoning about object semantics.
Models may overfit to contextual cues of the training set, resulting in poor generalization to unseen data, and in particular can lead to failure on long tail examples (\reffig{teaser}, \emph{ right}).
Out-of-context generalization is even more important in safety critical settings, such as autonomous driving, where rare but dangerous situations are often not well represented in training sets, \eg, pedestrians walking in the middle of the road or people appearing in the front of a vehicle (\reffig{teaser}, \emph{left}).


3D datasets are usually significantly smaller and exhibit little in-data variance compared to 2D image datasets that are often collected ``in the wild''\,\cite{lin14ECCV}.
High-capacity models \cite{Choy2019CVPR, Graham18CVPR, thomas19ICCV} are particularly vulnerable to learn strong context-based priors when trained on small 3D datasets.
As a result, modern network architectures for 3D processing often struggle with overfitting and typically expose large generalization gaps (\reffig{overfitting}).

To this end, we propose a new data augmentation technique, called Mix3D, that \emph{``mixes''} pairs of training scenes. Specifically, mixing consists in randomly transforming pairs of point clouds such that they loosely align, followed by taking the union over the two point sets (see \reffig{mix3d}).
This is implemented as concatenation of two point clouds as well as their corresponding training labels.
Mixing two scenes essentially generates new training samples, that increase the training set, and reduces overfitting as demonstrated in our analysis. 
In contrast to standard augmentations that apply transformations to single training samples without changing the scene context, Mix3D exposes the objects of each original scene to the novel context of the mixed scenes.
The result is an effective data augmentation strategy that can easily be incorporated into existing training pipelines, works well even without labeled data, and effectively reduces overfitting while increasing segmentation performance.

We evaluate Mix3D on three large-scale 3D semantic segmentation benchmarks:
ScanNet\,\cite{Dai17CVPR}, S3DIS\,\cite{Armeni16CVPR} and SemanticKITTI\,\cite{Behley19ICCV}.
We compare current state-of-the-art voxel-based MinkowskiNet\,\cite{Choy2019CVPR} and point-based KPConv\,\cite{thomas19ICCV} trained with and without Mix3D.
On all datasets and models, Mix3D consistently improves segmentation performance on both indoor and outdoor data.
Our analyses show that mixing scenes is key to consistent improvements and verify that Mix3D effectively reduces overfitting to the contextual priors of the training scenes.

\parag{Our contributions} are summarized as:

\begin{itemize}
\item We propose Mix3D, a data augmentation technique for large-scale 3D scenes and high-capacity models that can balance global context and local geometry.
As a result, the trained models can generalize beyond the contextual priors of the training set.
\vspace{-3px}
\item Applying Mix3D to recent point-based and voxel-based 3D models results in consistently improved segmentation performance on three large-scale indoor and outdoor datasets.
\vspace{-3px}
\item New state-of-the-art semantic segmentation performance of $78.1\%$ mean intersection over union (mIoU) on the ScanNet test benchmark challenge.
\vspace{-3px}
\item In-depth analyses of Mix3D show the importance of global context, local geometry and mixing scenes, in terms of model generalization and segmentation performance.
\end{itemize}


\section{Related Work}
\vspace{-15px}
\parag{Mixed Sample Data Augmentation.}
In contrast to standard augmentations applied to single samples, MixUp pioneered by Zhang \etal\,\cite{Zhang18ICLR} enforces smooth label transitions by linearly interpolating between pairs of input samples and their corresponding class labels\,\cite{Chen20ECCV, Guo19AAAI, Lee21CVPR, Verma19ICML, Yun19ICCV, Zhang18ICLR, Zhang21arxiv}.
Training on these virtual training samples encourages linear behavior outside the empirical train distribution, leading the classifier to show fewer label oscillations and, thus, improved generalization\,\cite{Zhang18ICLR}.
Numerous works extend MixUp:
AdaMixUp\,\cite{Guo19AAAI} takes special provisions to prevent manifold intrusions of virtual training examples.
Saliency guided approaches\,\cite{kim20icml, Uddin21ICLR} and (attentive) CutMix\,\cite{walawalkar20arxiv, Yun19ICCV} paste image crops into the image and thus increase localization accuracy.
In contrast to mixing train samples in the input domain, Verma \etal\,\cite{Verma19ICML} show that interpolating in the latent feature space leads to compressed representations which are theoretically linked to increased generalization capabilities.
However, these methods are inherently limited to dense representations such as images, waveforms and spectrograms\,\cite{Zhang18ICLR}, as they rely on one-to-one correspondences between parts of the input signals, \eg, pixel coordinates in an image.
Sparse representations such as point clouds do not provide readily accessible optimal bijective correspondences.
Therefore, PointMixUp\,\cite{Chen20ECCV} leverages the Earth Mover's Distance\,\cite{Rubner00IJCV} to interpolate between optimal pairwise assignments that lie on the shortest path between two point clouds.
Inspired by (Attentive) CutMix\,\cite{walawalkar20arxiv,Yun19ICCV} and saliency guided approaches\,\cite{kim20icml, Uddin21ICLR}, PointCutMix\,\cite{Zhang21arxiv} improves upon this work by replacing subsets with corresponding points of other point clouds.
Circumventing the problem of establishing bijective correspondences, Achituve \etal\,\cite{Achituve21WACV} explore combining point clouds for domain adaptation for single object semantic segmentation.
Analogously to Achituve \etal\,\cite{Achituve21WACV}, we treat augmented point clouds as their own entity, \ie, point features and associated labels do not change in the presence of a second point cloud.
Our approach is thus not limited by establishing optimal one-to-one point correspondences of equally sized sets.
Moreover, we do not rely on heuristic preprocessing such as aligning the main symmetry axes of point clouds\,\cite{Chen20ECCV}.
Thus, our approach is applicable to realistic tasks such as semantic segmentation of full scenes and is not limited to single objects of fixed point set sizes\,\cite{Chen20ECCV, Zhang21arxiv}.

\parag{Context Augmentations.}
Rosenfeld \etal\,\cite{rosenfeld18arxiv} show that object detection models are vulnerable towards placing out-of-context instances in an image and therefore over-rely on context information.
Shetty \etal\,\cite{shetty19cvpr} thus seek to alleviate the dependency to contextual variations by introducing a removal-based data augmentation where context or objects are cropped out of the image. 
Alternative methods propose to place novel instances into an image:
On the one hand, \cite{Alhaija18IJCV, dvornik18, tripathy18, Zhang20ECCV} constitute a line of work which places objects in the scene at visually \emph{plausible} locations.
Dvornik \etal\,\cite{dvornik18} show that merely placing additional instances in images in fact leads to decreased performances for 2D object detection.
However, by placing instances at semantically plausible positions in an image, they were able to report improved performances.
Similar to \cite{dvornik18}, Tripathi \etal\,\cite{tripathy18} propose an adversarial learning strategy that estimates plausible positions for additional instances.
Zhang \etal\,\cite{Zhang20ECCV} improve upon\,\cite{dvornik18, tripathy18} by conditioning the placement of instances not only on context information but also on the instance itself.
In the task of 3D semantic segmentation of outdoor scenes, Yan \etal\,\cite{yan18} carefully populate scenes with car instances taking special care to avoid overlapping objects. 
A number of other methods ignores the requirement for realistic data and randomly place object instances into the scene\,\cite{dwibedi17iccv, Ghiasi20arxiv} providing realistic context information only on a local patch-level.
This is comparable to \namelong{} as mixing two 3D scenes can also introduce unrealistic merging of objects.
In contrast to approaches adding instances into the input image, other approaches\,\cite{Chen2020GridMaskDA, Devries17arxiv, Singh2018HideandSeekAD, zhong2020erasing} explore augmentation methods for removing information from the input image.
The key idea is that rectangular regions of the input image are removed which reduces the risk to overfit to small train set specific context features\,\cite{Devries17arxiv}.
Unlike works aiming to retain a visually plausible context,
we argue that maintaining the context of augmented images amplifies the context bias.
We therefore seek to regularize its effects by adding instances together with their respective contexts and voluntarily create context overlaps.
Moreover, predicting plausible placements of objects demands complex augmentation systems, whereas our approach is fast, and can be added to existing approaches with a few lines of code (\cf \reffig{pipeline_code}, \emph{right}).
\vspace{-10pt}
\parag{Context Aggregation of Deep Learning Models.}
Point-based approaches such as\,\cite{Li18NIPS, Qi17CVPR, Qi17NIPS, thomas19ICCV, Wang18CVPRa, Wu18CVPR} perform semantic segmentation by splitting scenes into smaller chunks, effectively restricting the model's ability to learn from global context.
Subsequent works\,\cite{3dsemseg_ICCVW17, Engelmann20ICRA, huang2018recurrent, ye20183d} report improvements by increasing the spatial context.
By leveraging data efficient sparse convolutions, recent voxel-based methods \eg, MinkowskiNet\,\cite{Choy2019CVPR} and SparseConvNet\,\cite{Graham18CVPR} are capable of processing full scenes at once, thus, capturing the global context of the scene.
3D-MPA\,\cite{Engelmann20CVPR} and DOPS\,\cite{ross2020dops} extend on this idea for instance segmentation and object detection.
They learn higher-level relationships among object proposals with the goal to learn global scene configuration rules\,\cite{oliva07cogn}.
Context is equally important in recent methods for 3D reconstruction \cite{engelmann2020points,popov2020corenet}.
These developments allow to fully capture and rely on global scene context.
However, our experiments show that this can have detrimental implications.
By applying \namelong{}, we seek to generalize beyond the contextual priors of the training set.


\section{Method}
\label{sec:method}
\vspace{-5px}
We present a data augmentation technique for 3D deep learning on realistic large-scale scenes.
The goal is to train 3D models that are less biased by misleading context priors, and learn to balance between local structures and global scene context without overfitting to training scenes.

\begin{figure}[t]
     \centering
\includegraphics[width=\columnwidth]{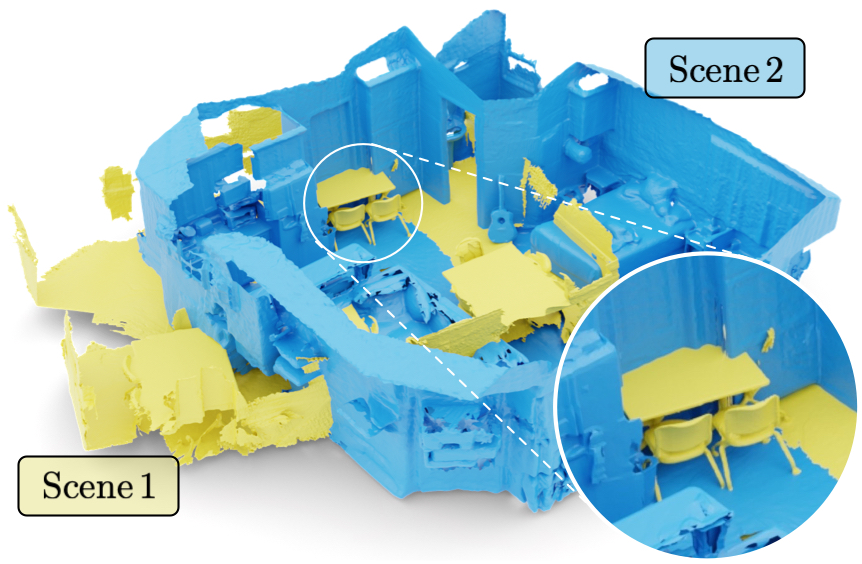}
     \caption{
     \small
     \textbf{Illustration of two mixed 3D scenes.}
     The Mix3D data augmentation consists in taking the union over two 3D scenes after random transformations while ensuring sufficient overlap.
     This results in the effect of objects appearing now in the novel context of both mixed scenes, \eg, the classroom table directly faces the apartment door.
     The goal is to train a model that learns when to rely on context and when to focus on local geometry.}
     \label{fig:mix3d}
     \vspace{-5px}
 \end{figure}

\begin{figure*}[t]
  \vspace{10px}
     \centering
     \includegraphics[width=0.66\textwidth]{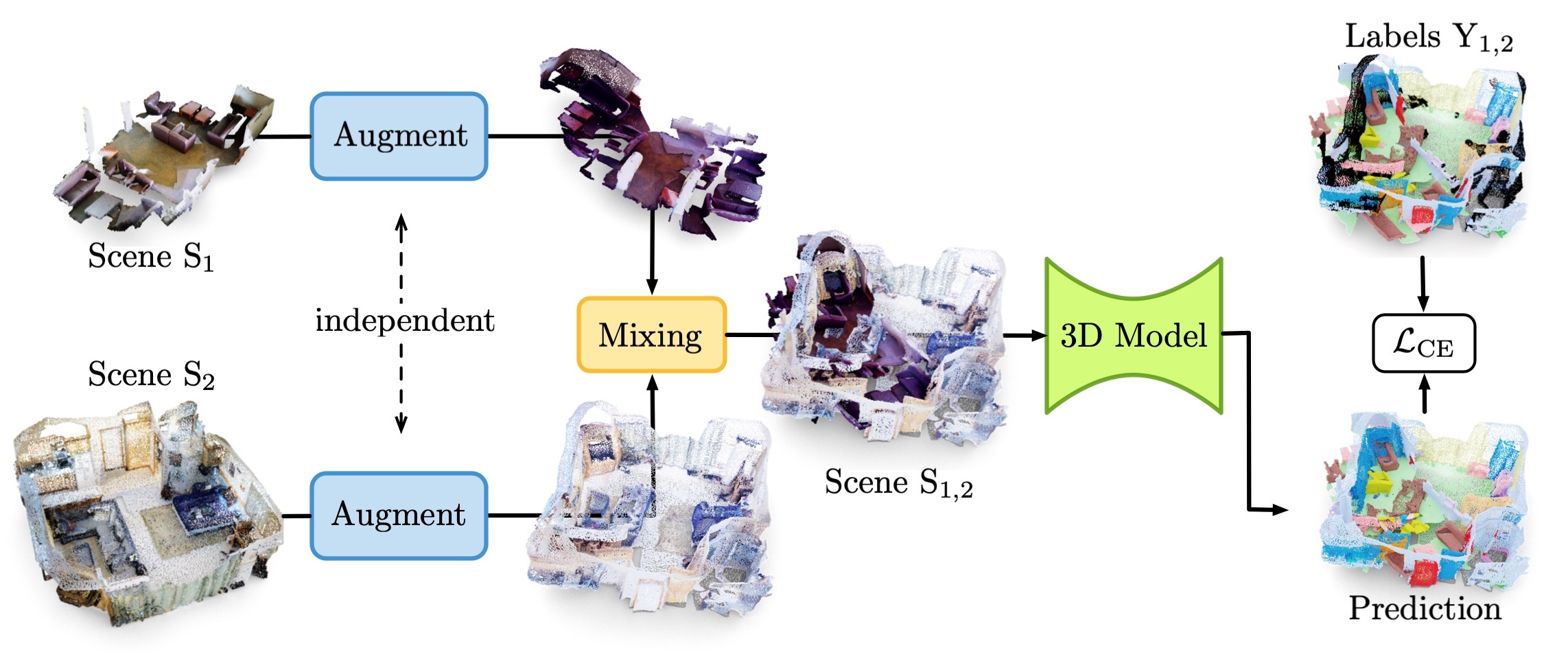}
     \includegraphics[width=0.32\textwidth]{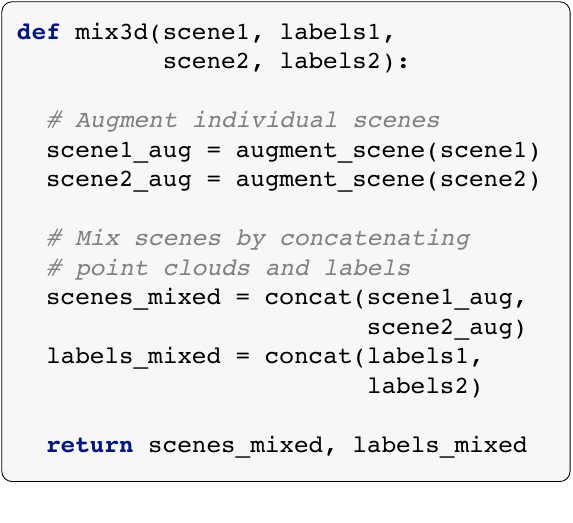} 
     \caption{\small
     \emph{(Left)}
     \textbf{Training pipeline.} Mix3D is easy to incorporate into existing code bases: instead of feeding a single scene S$_1$ into the 3D model, a second input scene S$_2$ is augmented in parallel and mixed with the first one.
     The resulting mixed scene S$_{1,2}$ is input to the 3D model, which remains unchanged.
     For semantic segmentation, we compute the standard cross entropy loss $\mathcal{L}_{\text{CE}}$ on the predicted labels and the concatenated ground truth labels Y$_{1,2}$ of both scenes.
     \emph{(Right)}
          \textbf{Exemplary implementation of Mix3D.}
     The individual scenes are augmented by centering at the origin, followed by random rotation and translation while ensuring overlap between both scenes.
     Additionally, we perform augmentations such as color-jitter, elastic distortion etc. (Details in \refsec{method}).
     Finally, since the order of the points does not change during augmentation, the ground truth semantic labels are obtained using concatenation as well.}
     \label{fig:pipeline_code}
 \end{figure*}

Mix3D creates novel training examples by mixing two original scenes.
The mixing of 3D scenes is illustrated in \reffig{mix3d}.
In particular, we expose objects from a single input scene to the combined context of both mixed scenes.
By doing so, the network does not only need to learn how to implicitly disentangle two mixed scene contexts, but also sees every object in a large variety of object arrangements that one would not normally encounter.
Most notable, from initially $N$ different scene contexts, with Mix3D we obtain $\mathcal{O}(N^2)$ novel contexts via pairwise combination of the existing scenes. 
In the following, we explain the details of the Mix3D training pipeline (\reffig{pipeline_code}, \emph{left}).

\parag{Data Augmentation} (\reffig{pipeline_code}, \colorsquare{m_blue}).
When augmenting input scenes, we first  translate them to the origin by subtracting the centroid from all point positions.
By doing so, we make sure that the two scenes are overlapping in the following mixing stage.
We randomly flip the point cloud in both horizontal directions, randomly rotate the scene along the up-right axis and along the other axis by Uniform$[-\frac{\pi}{64}, +\frac{\pi}{64}]$.
We also apply random sub-sampling, elastic distortion\,\cite{Choy2019CVPR,ronneberger2015u} and randomly scale the scene by Uniform$[0.9, 1.1]$. When the input features include color, we additionally apply random brightness and contrast augmentation as well as color-jitter. 

\parag{Mixing} (\reffig{pipeline_code}, \colorsquare{m_orange}).
After augmenting, both scenes are mixed together, which is simply the union of both augmented point clouds.
Technically, we implemented this by concatenating batch entries in pairs.
Moreover, since the order of the points is not modified during augmentation,
the ground truth labels of the mixed point cloud are also obtained by concatenation (\reffig{pipeline_code}, \emph{right}).

\parag{3D Model} (\reffig{pipeline_code}, \colorsquare{m_green}).
Mix3D is independent of the underlying 3D deep learning model.
In experiments, we use MinkowskiNet\,\cite{Choy2019CVPR} and KPConv\,\cite{thomas19ICCV}
as state-of-the-art representatives of both voxel-based and point-based models.
The exact experimental setup is described in more detail in \refsec{experiments} and the supplementary.

\parag{Loss function.} We supervise the model for semantic segmentation using the standard cross-entropy loss $\mathcal{L}_{\text{CE}}$ which we apply to the predictions of the mixed scene S$_{1,2}$ and the concatenated ground truth labels Y$_{1,2}$ of both scenes.

\parag{Discussion.} Crucially, \namelong{} differs from existing point cloud mixing-based techniques in two main aspects.
Firstly, we are the first to explore mixed sample augmentations beyond single object classification for tackling large-scale scene segmentation.
We see this as an important step towards improving real-world 3D scene understanding.
Second, inspired by MixUp\,\cite{Zhang18ICLR}, recent techniques propose to interpolate labels\,\cite{Lee21CVPR, Zhang21arxiv} and even input samples\,\cite{Chen20ECCV}.
While appropriate for a holistic classification task, here we keep the mixed samples and labels in their original form.
We motivate this in the following manner:
consider the task of predicting the convex combination of two inputs $(aS_1,(1-a)S_2)$ as $aY_1 + (1-a)Y_2$.
This implies that the prediction of $Y_1$ changes in the presence of $Y_2$.
Yet, this is the opposite of our aim here, where we wish to keep the prediction of $Y_1$ agnostic to the existence of $Y_2$.
In particular, we preserve the complete context information of each mixed sample, whereas PointMixUp\,\cite{Chen20ECCV} distorts point clouds by interpolation and RSMix\,\cite{Lee21CVPR} only preserves the spatial structure of locally restricted chunks.
We provide pseudo code in Figure\,\ref{fig:pipeline_code} \emph{(right)}.

\section{Experiments}
\label{sec:experiments}
In this section, we first compare a variety of state-of-the-art models trained with and without Mix3D on three large-scale indoor and outdoor 3D semantic segmentation benchmarks (\refsec{sota}).
We provide detailed analysis experiments to motivate and understand the importance of scene context, local geometry and mixing scenes (\refsec{analysis}).
Finally, we show additional qualitative results.
More analysis and results are found in the supplementary.

\subsection{Comparison with State-of-the-art Methods}
\label{sec:sota}
\begin{table*}
\centering
        \begin{tabular}{r c c c c c}
            \toprule
            &&
            \hspace{5mm} \textbf{ScanNet}\,\cite{Dai17CVPR} \hspace{5mm} &
            \hspace{5mm} \textbf{S3DIS}\,\cite{Armeni16CVPR} \hspace{5mm} &
            \multicolumn{2}{c}{
            \hspace{5mm} \textbf{SemanticKITTI}\,\cite{Behley19ICCV}\hspace{5mm} \,}\\
             Model & \name{} & Validation &  Area-5 & Validation & Test \\
            \midrule
                \multirow{2}{*}{MinkowskiNet~\cite{Choy2019CVPR}}  & 
                \xmark & 
                $72.4$ {\footnotesize $\pm$ $0.18$} & 
                $64.7$ {\footnotesize $\pm$ $0.34$} &
                $56.7$ & $53.2$ \\
                & \cmark &
                $\mathbf{73.6}$ {\footnotesize $\pm$ $0.31$} & 
                $\mathbf{65.4}$ {\footnotesize $\pm$ $0.37$} &
                $\mathbf{59.9}$ & $\mathbf{58.1}$ \\ 
            \midrule
                \multirow{2}{*}{Rigid KPConv~\cite{thomas19ICCV}} & 
                \xmark & 
                \expBaselineScannetKPConvRigid &
                \expBaselineStanfordKPConvRigid & $65.1$ & $63.1$ \\
                 & \cmark & 
                 \expMergedScannetKPConvRigid & \expMergedStanfordKPConvRigid & $\mathbf{66.6}$ & $\mathbf{63.6}$ \\
            \midrule
                \multirow{2}{*}{Deformable KPConv~\cite{thomas19ICCV}} &
                \xmark & 
                \expBaselineScannetKPConvDeform &
                \expBaselineStanfordKPConvDeform & $-$ & $-$\\
                & \cmark & 
                \expMergedScannetKPConvDeform & 
                \expMergedStanfordKPConvDeform & $-$ & $-$ \\
            \bottomrule
    \end{tabular}
\vspace{-3pt}
\caption{\small \textbf{Effect of Mix3D on models for 3D semantic segmentation.}
We compare MinkowskiNet (voxel-based) and KPConv (point-based), with and without Mix3D on large-scale indoor scenes (ScanNet, S3DIS) and outdoor scenes (SemanticKITTI).}
\label{tab:sota_comparison}
\end{table*}
\begin{table*}
\centering
\resizebox{\textwidth}{!}{%
\setlength{\tabcolsep}{1.8pt}
\begin{tabular}{rcc|cccccccccccccccccccc}
        \toprule
        Method & \namelong{} & mIoU & wall & floor & cabinet & bed & chair & sofa & table & door & wndw & books. & pic & counter & desk & curtain &\cellcolor{LightCyan} fridge & shower & toilet & sink & bathtub & otherf.\\
        \midrule
        MinkNet\,\cite{Choy2019CVPR} & \xmark & 72.4 & 85.6 & 96.5 & 64.7 & \textbf{82.1} & 91.0 & 84.4  & 74.5 & \textbf{65.0} & 62.8 & 79.5 & 32.4 & 64.4 & 63.7  & \textbf{75.5} & \cellcolor{LightCyan}51.6  &  \textbf{69.0}  & 93.0  & 67.6 & 87.6  & \textbf{56.3} \\
        (2cm) & \cmark & \textbf{73.6}  & \textbf{86.0}  & \textbf{96.6} & \textbf{66.3} & \textbf{82.1} & \textbf{91.9} & \textbf{86.1} & \textbf{75.7}  & 64.7 & \textbf{64.7} & \textbf{79.6} & \textbf{36.3} & \textbf{67.7} & \textbf{67.0} & 74.7 &\cellcolor{LightCyan} \textbf{59.3} & 68.8 & \textbf{93.9} & \textbf{68.6} & \textbf{87.9} & 55.1 \\
        \midrule
        KPConv\,\cite{thomas19ICCV} & \xmark & 69.3 & 82.4 & \textbf{94.4} & 64.5 & 79.2 & 88.5 & 77.2 & \textbf{73.0} & 60.5 & \textbf{59.1} & \textbf{79.8} & 28.4 & 59.9 & \textbf{63.7} & 71.6 & \cellcolor{LightCyan}53.1 & 54.1 & 91.5 & 63.3 & \textbf{86.0} & \textbf{56.4} \\
        (Deformable) & \cmark & \textbf{70.3} & \textbf{82.6} & \textbf{94.4} & \textbf{65.8} & \textbf{79.5} & \textbf{90.2} & \textbf{79.9} & 72.5 & \textbf{61.7} & 58.2 & 79.6 & \textbf{31.5} & \textbf{62.1} & 63.2 & \textbf{72.0} & \cellcolor{LightCyan}\textbf{58.1} & \textbf{56.8} & \textbf{91.8} & \textbf{64.6} & 85.6 & 55.4 \\
        \bottomrule
\end{tabular}%
}
\vspace{-3pt}
\caption{\small \label{tab:scannet_val_scores} \textbf{Semantic segmentation scores on ScanNet validation\,\cite{Dai17CVPR}.}
We report the mean per-class scores over three trained models.
}
\label{tab:scannet_per_class}
\end{table*}

\parag{Datasets.}
S3DIS\,\cite{Armeni16CVPR} consists of dense 3D point clouds from 6 large-scale areas including 271 rooms from 3 different buildings.
The annotations cover 13 semantic classes. We follow the common evaluation protocol \cite{Qi17CVPR, Tchapmi173DV, Wang18CVPRa} and train on all areas except Area~5, which serves as test-set.
ScanNet\,V2\,\cite{Dai17CVPR} contains 3D scenes of a wide variety of indoor rooms annotated with 20 semantic classes.
We follow the public training, validation, and test split of 1201, 312 and 100 scans, respectively.
SemanticKITTI\,\cite{Behley19ICCV} shows large-scale outdoor traffic scenes recorded with a Velodyne-64 laser scanner. 
It consists of $19130$ training scenes, recorded over $10$ different driving sequences, and a validation set of $4071$ samples.
There are $28$ annotated semantic classes from which $19$ are evaluated.
We report scores in the single scan setting, \ie, individually per frame.

\parag{Models in comparison.}
We apply Mix3D to MinkowskiNet\,\cite{Choy2019CVPR} and KPConv\,\cite{thomas19ICCV}, as the recent best-performing point-based and voxel-based methods.
MinkowskiNet\,\cite{Choy2019CVPR} is a voxel-based model that implements the idea of sparse convolutional networks originally presented by Graham \etal\,\cite{Graham18CVPR}.
KPConv\,\cite{thomas19ICCV} represent the currently best performing point-based approach.
Its kernel weights are located in Euclidean space which is more flexible than fixed grid positions.
The kernel point positions are either deformable or defined on a rigid pattern.
In our experiments, we evaluate both rigid and deformable KPConvs.
For all models, we use the official code.
For MinkowskiNets on S3DIS and SemanticKITTI, we use our own training pipeline.

\parag{Results} are presented in \reftab{sota_comparison} and \reftab{scannet_per_class}.
For each model, we report the mean (and standard deviation) over three training runs.
Training with Mix3D consistently improves previous state-of-the-art models (MinkowskiNet\,\cite{Choy2019CVPR}, KPConv\,\cite{thomas19ICCV}) across 3 different indoor and outdoor datasets by a statistically significant margin.
The most notable improvement is obtained on SemanticKITTI with up to $+4.9$\,mIoU.
Importantly, these results are obtained by only adding Mix3D, while keeping training hyper-parameters unchanged, including training time and effective batch size (same number of points per batch) for all datasets.
\reftab{scannet_per_class} shows that the class ``fridge", with one of the fewest training samples, improves significantly by $+5$\,mIoU with KPConv and $+7.7$\,mIoU (MinkowskiNet).
We also train a MinkowskiNet ensemble with Mix3D and report state-of-the-art scores on the ScanNet benchmark challenge\,\cite{Dai17CVPR} of $\mathbf{78.1}$\,\textbf{mIoU} (\reftab{scannet_benchmark}).
Each model is trained for 120k iterations, \ie, the same number of training iterations as suggested by Choy \etal\,\cite{Choy2019CVPR}, on both the train and validation split, and we follow \cite{Graham18CVPR,Schult20CVPR,thomas19ICCV} for test time augmentation.
More architectural and training details, as well as per-class results on all datasets are provided in the supplementary.

\vfill

\begin{table}[b]
    \centering
    \resizebox{\columnwidth}{!}{%
    \begin{tabular}{l c c c}
        \toprule
         \multicolumn{4}{c}{\textbf{ScanNet Semantic Segmentation}} \\
         \multicolumn{4}{c}{Benchmark Challenge (hidden test set)} \\
        \toprule
        Method 								&							& Input 			& mIoU \\
        \midrule
		\rowcolor{LightCyan}Mix3D (ours)  					& 3DV'21							& 3D 			& $\mathbf{78.1}$ \\
		OccuSeg\,\cite{Han20CVPR} & CVPR'20						& 3D 			& $76.4$ \\
		O-Net\,\cite{Wang-2017-ocnn} & SIGGRAPH'17 & 3D & $76.2$ \\
		BPNet\,\cite{Hu21CVPR}		&CVPR'21						& 3D + 2D 	& $74.9$ \\
		VMNet\,\cite{hu2021vmnet} & ICCV'21 & 3D + Mesh & $74.6$ \\
		Virtual MVFusion\,\cite{Kundu20ECCV}& ECCV'20			& 3D + 2D		& $74.6$ \\
		MinkowskiNet\,\cite{Choy2019CVPR} &CVPR'19			& 3D				& $73.6$ \\
		SparseConvNet\,\cite{Graham18CVPR} &CVPR'18		& 3D				& $72.5$ \\
		RFCR\,\cite{Gong21CVPR}	& CVPR'21						&	3D			& 	$70.2$ \\
		JSENet\,\cite{Hu20ECCV}	& ECCV'20							&	3D			& 	$69.9$ \\
		FusionNet\,\cite{Zhang20ECCVb}	& ECCV'20				& 3D				& 	$68.8$ \\
		KP-Conv\,\cite{thomas19ICCV} & ICCV'19					&	3D			& 	$68.4$ \\
		PointConv\,\cite{Wu18CVPR}& CVPR'18						&	3D			& 	$66.6$ \\
		PointASNL\,\cite{Yan20CVPR}	& CVPR'20					&	3D			& 	$66.6$ \\
		DCM-Net\,\cite{Schult20CVPR}& CVPR'20					&	3D + Mesh& 	$65.8$ \\
        \bottomrule
    \end{tabular}
}
\vspace{-6pt}
\caption{
\small \textbf{Semantic segmentation results on ScanNet test.}
We include methods that additionally rely on 2D images and polygon meshes as input.
Benchmark accessed on 5\textsuperscript{th} October 2021.}
\vspace{-6pt}
\label{tab:scannet_benchmark}
\end{table}

\newpage

\subsection{Motivation and Analysis Experiments}
\label{sec:analysis}

In this section, we experimentally motivate and analyze Mix3D.
Unless stated otherwise, the model is MinkowskiNet ($5$\,cm voxels) trained on the ScanNet train split and evaluated on the validation split for semantic segmentation.

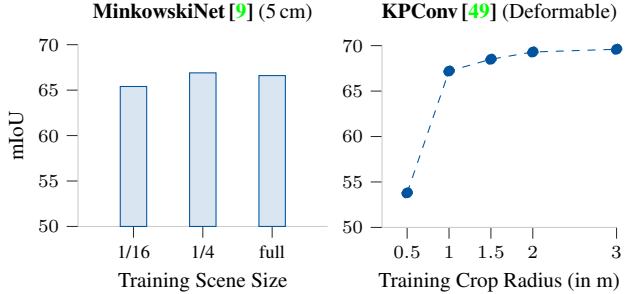
\begin{figure}[t]
\begin{footnotesize}
\begin{tikzpicture}
\begin{axis}[width=4.9cm, height=4cm, xtick=data,
	axis x line=bottom,
axis y line=left,
x axis line style={-},
y axis line style={-},
axis line style={white!80!black},
legend cell align={left},
legend style={draw=none, fill opacity=0.8, text opacity=1},
tick align=outside,
x grid style={white!80!black},
ymin=50, ymax=70,
ytick style={color=white!15!black},
xtick style={color=white!15!black},
tick label style={font=\scriptsize, xshift=0},
y grid style={white!80!black},
ybar,
enlarge x limits=0.4,
symbolic x coords={1/16,1/4, full},
title={\textbf{MinkowskiNet\,\cite{Choy2019CVPR}} (5\,cm)},
		xlabel={Training Scene Size},
		ylabel={mIoU},
		ytick style={color=white!15!black}
		],

\addplot[color=palette_sinopia, fill=palette_sinopia!15,
error bars/.cd, y dir=both, y explicit] coordinates {
(1/16, 65.4)
(1/4, 66.9)
(full, 66.6)};
\end{axis}
\end{tikzpicture}
\begin{tikzpicture}
	\begin{axis}[width=4.7cm, height=4cm, xtick=data,
	axis x line=bottom,
axis y line=left,
x axis line style={-},
y axis line style={-},
axis line style={white!80!black},
legend cell align={left},
legend style={draw=none, fill opacity=0.8, text opacity=1},
tick align=outside,
x grid style={white!80!black},
ymin=50, ymax=70,
xmin=0.2,
ytick style={color=white!15!black},
xtick style={color=white!15!black},
tick label style={font=\scriptsize, xshift=0},
y grid style={white!80!black},
title={\textbf{KPConv\,\cite{thomas19ICCV}} (Deformable)},
		xlabel={Training Crop Radius (in m)},
		ylabel={},
		ytick style={color=white!15!black}
		],
	\addplot[color=palette_sinopia,dashed,mark=*]
	coordinates {
		(0.5,53.8)
		(1.0,67.2)
		(1.5,68.5)
		(2.0,69.3)
		(3.0,69.6)};
	\end{axis}
\end{tikzpicture}
\end{footnotesize}
\vspace{-15px}
\caption{\textbf{Influence of context during training.}
Increasing context during training improves semantic segmentation performance on ScanNet validation.
However only up to a certain point at which overfitting to scene context limits performance. 
}
\label{fig:context}
\end{figure}

\parag{What is the effect of context?}
Recent developments focus on models with ever larger receptive fields, such that they can capture larger scene context.
The intuition is that global context provides additional information that is helpful for making local decisions.
To verify this hypothesis, we artificially reduce the scene context. In particular,  we train different models on smaller regions of a scene, \ie, on crops of smaller sizes.
The key idea is that, during training, the model sees only a fraction of the original scene, and therefore it cannot rely on global scene context (only on reduced context within the crop). 
In this experiment, we train MinkowskiNet\,\cite{Choy2019CVPR} models on increasing chunk sizes,
as well as KPConv\,\cite{thomas19ICCV} models on spheric crops of increasing radius.
Indeed, the results in \reffig{context} show that increasing the crop size (and thus the context) is in general helpful.
However, we also observe that MinkowskiNet models trained on $\sfrac{1}{4}$-fractions of a scene perform on-par to models trained on full scenes.
This suggests that relying too much on global context can reduce performance.
A similar effect is observed for KPConv where the performance does not significantly increase after $2$\,m radius.
In conclusion, global context does indeed improve performance, however only up to a certain point at which overfitting to context limits the performance.

\parag{Does Mix3D help to focus more on local geometry?} %
In the previous experiment, we have seen that increased global context is not necessarily beneficial as overfitting to scene priors can occur.
The goal of the next experiment is to demonstrate that models trained with Mix3D
are less reliant on context information and
are able to make better use of local geometry and structures. 
For this experiment, the setup during training remains unchanged, \ie, we train the model with full scenes from the ScanNet training split.
At test time, however, we simulate missing context by showing only isolated objects to the trained network.
The individual objects are cropped out using the object mask annotations available in the ScanNet validation split.
This experiment shows how much the model depends on context, and how much it can make use of local geometry. 
In \reftab{val_on_instances}, we show semantic segmentation scores for \emph{individual} instances (mean IoU) for two types of models, one trained with, and one trained without Mix3D.
%

\begin{table}[h]
    \centering
    \begin{tabular}{l c c}
    \toprule
         			& \multicolumn{2}{c}{mIoU}\\
        Method 	& \multicolumn{2}{c}{\small(on single instances)} \\
    \midrule
        MinkowskiNet\,\cite{Choy2019CVPR} & \expInstanceValBase & \multirow{2}{*}{\ArrowDown{$+11.5$}} \hspace{-20px}\\
        MinkowskiNet + Mix3D\,{\small(ours)} & \textbf{\expInstanceValMerged} \\
    \bottomrule
    \end{tabular}
    \vspace{-5px}
    \caption{\small \textbf{Evaluation on single validation instances.}
    Using the ground truth instance masks, we evaluate on cropped out single instances.
    Both models are trained on full scenes.
    The Mix3D model improves over the model trained without \namelong{} by $+11.5$ mIoU.
    We report mean IoU scores and standard deviations of three trained models each.
    }
    \vspace{-1pt}
    \label{tab:val_on_instances}
\end{table}
While the overall scores are significantly lower than in full-scene validation experiments -- showing the importance of global context -- we also see that the Mix3D models perform significantly better ($+11.5$\,mIoU) on single instances than models trained without Mix3D.
In particular, this means that Mix3D models are notably less dependent on global context to make more accurate predictions given only local geometry. As such, we argue that they are less prone to overfitting on strong context priors and more importantly, Mix3D models can extract more discriminative information from local geometry alone.
We conclude: although exposed to larger crops, models trained without Mix3D are limited in their ability to leverage context information fully (\cf \reffig{context}). As seen in this experiment, Mix3D trained models, however, balance better between global context and local geometry and are thus less prone to overfit to context cues, resulting in better performance.

\parag{Do alternative out-of-context augmentations work?}
We have already seen that Mix3D influences the scene context during training so that the model overfits less on global context and relies more on local geometry by exposing objects of one scene to the context of another scene, and vice versa.
Next, we want to see if it is possible to find a simpler approach that achieves a similar effect as Mix3D.
First, we look at random noise as a means of obscuring scene context as suggested in \cite{Chapelle01NIPS}.
We compare two different noise patterns:
first, we add random points near the surface of 20\% of original points, \ie, within a radius of $0.5$\,m around the original point.
The intuition is that this noise pattern only affects instances locally by obscuring their context.
The second approach consists in adding random, uniformly distributed points within the bounding volume of the scene, which can enable non-local information flow.
For each 60$\text{cm}^3$ cube we add a point with a $0.1$\,m random offset.
\reftab{noise}, \circlenum{1} shows the resulting scores: adding these random noise patterns drastically reduces the performance by more than $6.5\%$ mIoU compared to the baseline.
Next, we adapt CutOut\,\cite{Devries17arxiv} from the image domain and remove points that fall into randomly sampled cubical chunks.
This resembles a \emph{thinning-out} effect \cite{srivastava2014dropout} on the context level which reduces the risk of overfitting to small training set-specific context features\,\cite{Devries17arxiv}.
We evaluate varying cuboid sizes and cutting frequencies (see \reftab{noise}, \circlenum{2}).
The experiments show that CutOut is sensitive to the parameter choice and that Mix3D still outperforms the best performing CutOut model. 
Last, we mix a second scene without ground truth labels, \ie, the semantic segmentation loss is computed only for points from one scene. 
While this model (\reftab{noise}, \circlenum{3}) outperforms the baseline significantly, it does not reach the performance of the fully supervised model. Nevertheless, this reveals an additional strength of Mix3D: the ability to use unlabeled scenes for mixing demands minimal acquisition effort since no human effort for labelling is required.

\begin{table}[h]
\centering
\begin{footnotesize}
\begin{overpic}[unit=1mm,width=8.1cm]{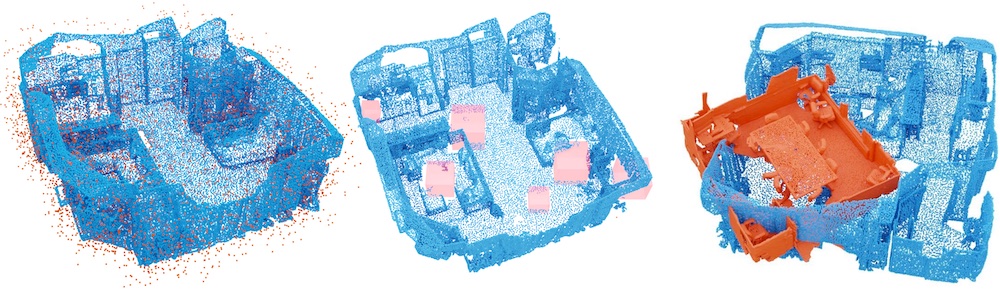}
\put(10, -2){\circlenum{\tiny 1} Noise}
\put(40, -2){\circlenum{\tiny 2} CutOut}
\put(75, -2){\circlenum{\tiny 3} Mixing}
\end{overpic}
\end{footnotesize}
\\
\vspace{10px}
\begin{small}
\begin{tabular}{l c c c}
\toprule
Augmentation &  && mIoU \\
\midrule
\multicolumn{2}{l}{Baseline} && $66.6$ \\
\midrule
\rowcolor{LightCyan}
\circlenum{1} Noise & Setting & Frequency & \\
 & near surface & 20\% & \expNoise \\
 & uniform & $1/60\,\text{cm}^3$ & $60.1$ \\
\midrule
\rowcolor{LightCyan}
\circlenum{2} CutOut  & Size & Cut frequency &  \\
\rowcolor{LightCyan}
        & {\small in meters} & {\small \#cuts/$10^4$points} &  \\
        &0.05$-$2.0 & 1 & \expCutoutAA \\
        &0.05$-$2.0 & 2 & \expCutoutBA \\
        &0.05$-$1.0 & 5 & \expCutoutCB \\
        &0.05$-$1.0 & 10 & \expCutoutDB \\
        &0.05$-$0.5 & 20 & \expCutoutEC \\
        &0.05$-$0.5 & 33 & \expCutoutFC \\
\midrule
\rowcolor{LightCyan}
\circlenum{3} Mixing &  & & \\
&\multicolumn{2}{l}{ without labels} & \expNoiseStructured \\
&\multicolumn{2}{l}{ with labels (\ie Mix3D)} & $\mathbf{69.0}$ \\
\bottomrule
\end{tabular}
\end{small}
\vspace{-5px}
\caption{
\small
\textbf{Alternative Context Changes.}
We compare random noise patterns \circlenum{1}, cutting out random chunks \circlenum{2} and mixing scenes with and without annotated labels \circlenum{3}. Mix3D with labels performs best, while mixing without labels is still a viable approach when large amounts of unlabeled data are available or too costly to label.
}
\label{tab:noise}
\end{table}

\definecolor{bike}{rgb}{0.3922, 0.90196, 0.96078}
\definecolor{car}{rgb}{0.3922, 0.5882, 0.96078}
\definecolor{other}{rgb}{1.0,0.4706,0.1961}
\definecolor{otherfurn}{rgb}{0.32156863, 0.32941176, 0.63921569}
\definecolor{refrigerator}{rgb}{1.        , 0.49803922, 0.05490196}
\definecolor{window}{rgb}{0.3922, 0.3922, 1.0000}

\begin{figure*}
\vspace{-5pt}
\begin{center}
\rotatebox{90}{\footnotesize \ \ \ \ \ \ \ \ \  Baseline Aug. + \namelong{} \ \ \ \ \ \ \ Baseline Augmentation}
\begin{subfigure}[b]{0.32\textwidth}
    \centering
    \includegraphics[width=\textwidth, trim={0 0 0 0}, clip]{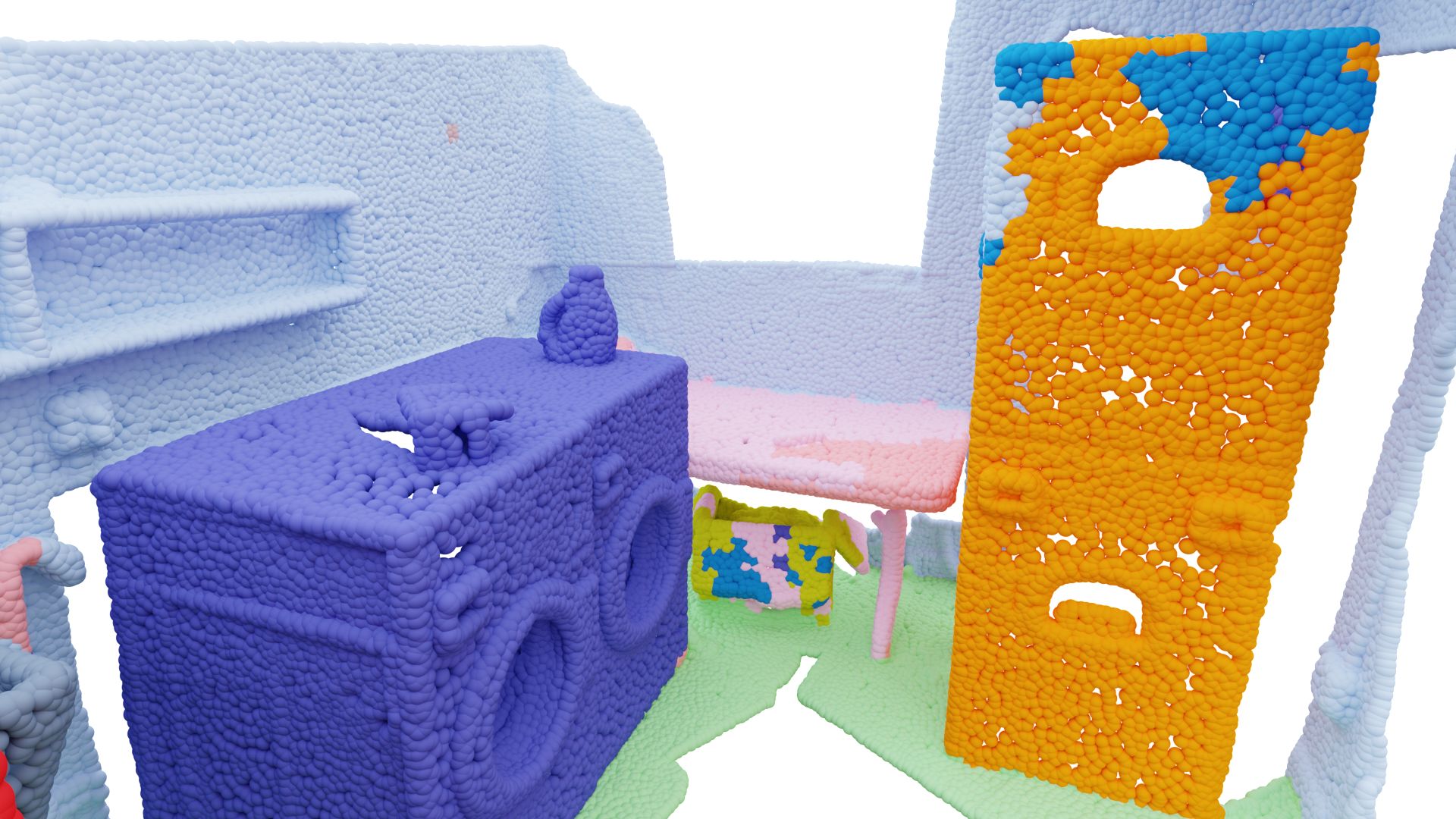}\\
    \includegraphics[width=\textwidth, trim={0 0 0 0}, clip]{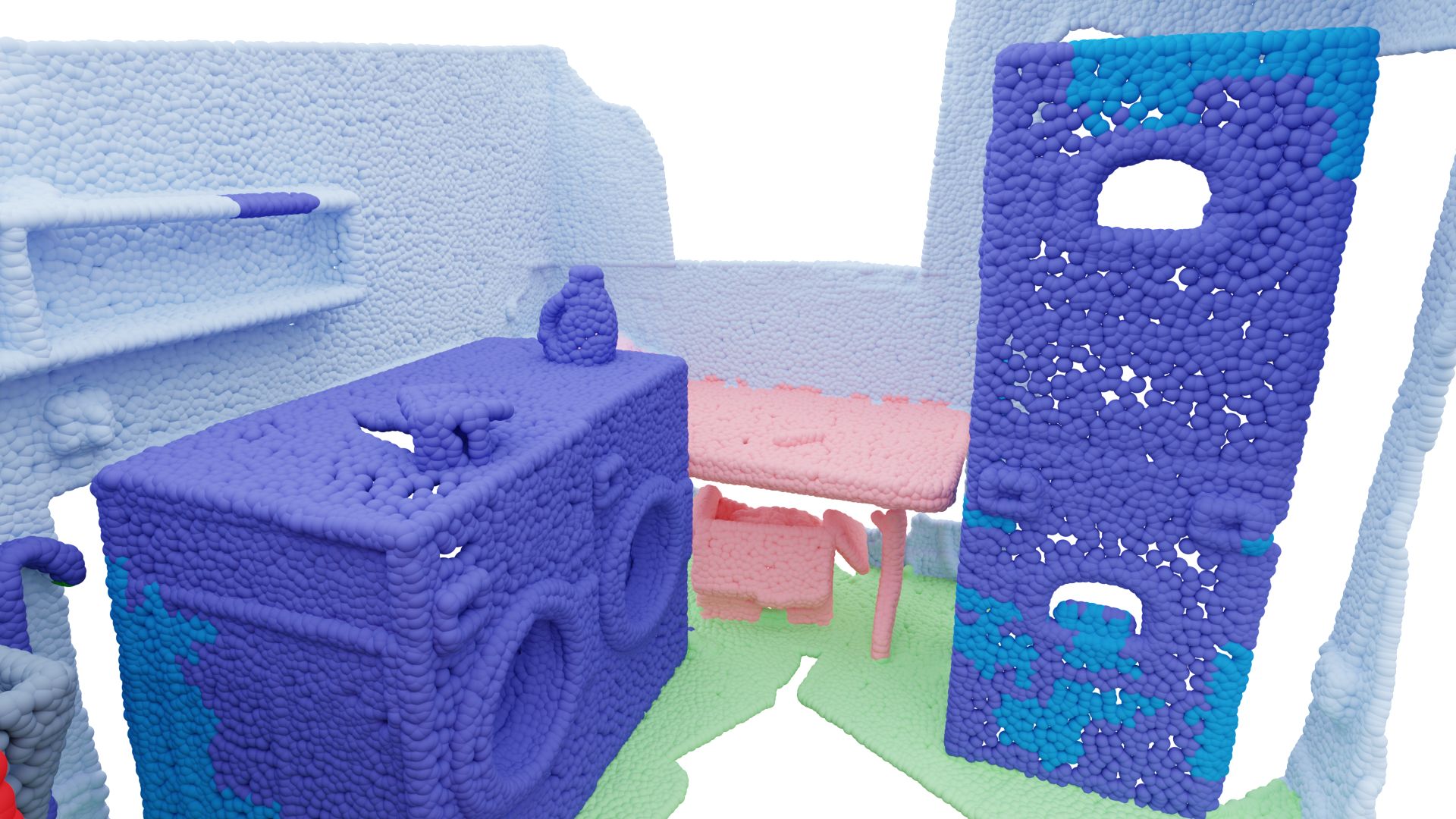}\\
    \vspace{-3px}
    \caption{\label{fig:semkitti} ScanNet\,\cite{Dai17CVPR}}
\end{subfigure}
\hfill
\begin{subfigure}[b]{0.32\textwidth}
    \centering
        \includegraphics[width=\textwidth, trim={0 0 0 0}, clip]{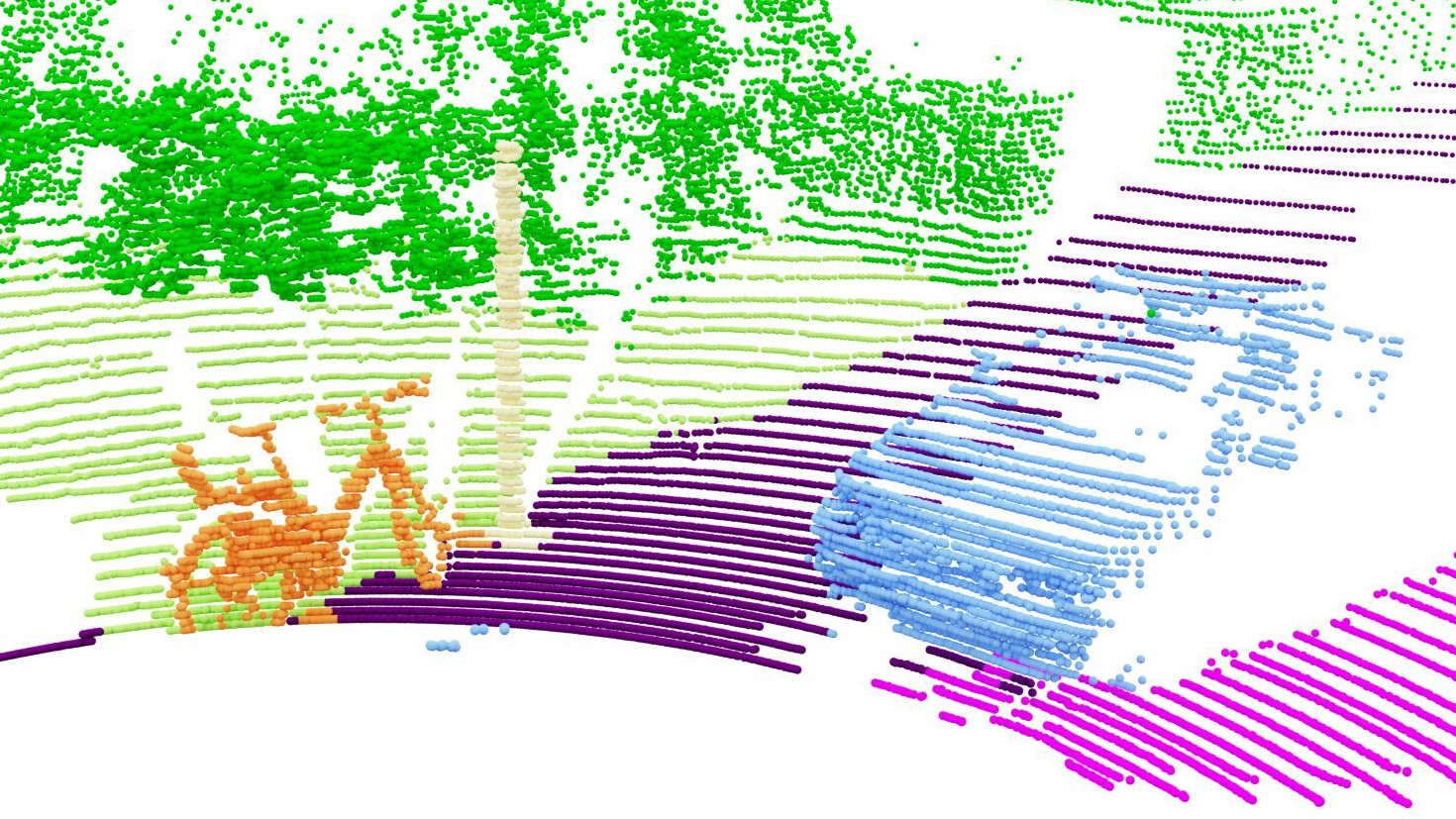}\\
    \includegraphics[width=\textwidth, trim={0 0 0 0}, clip]{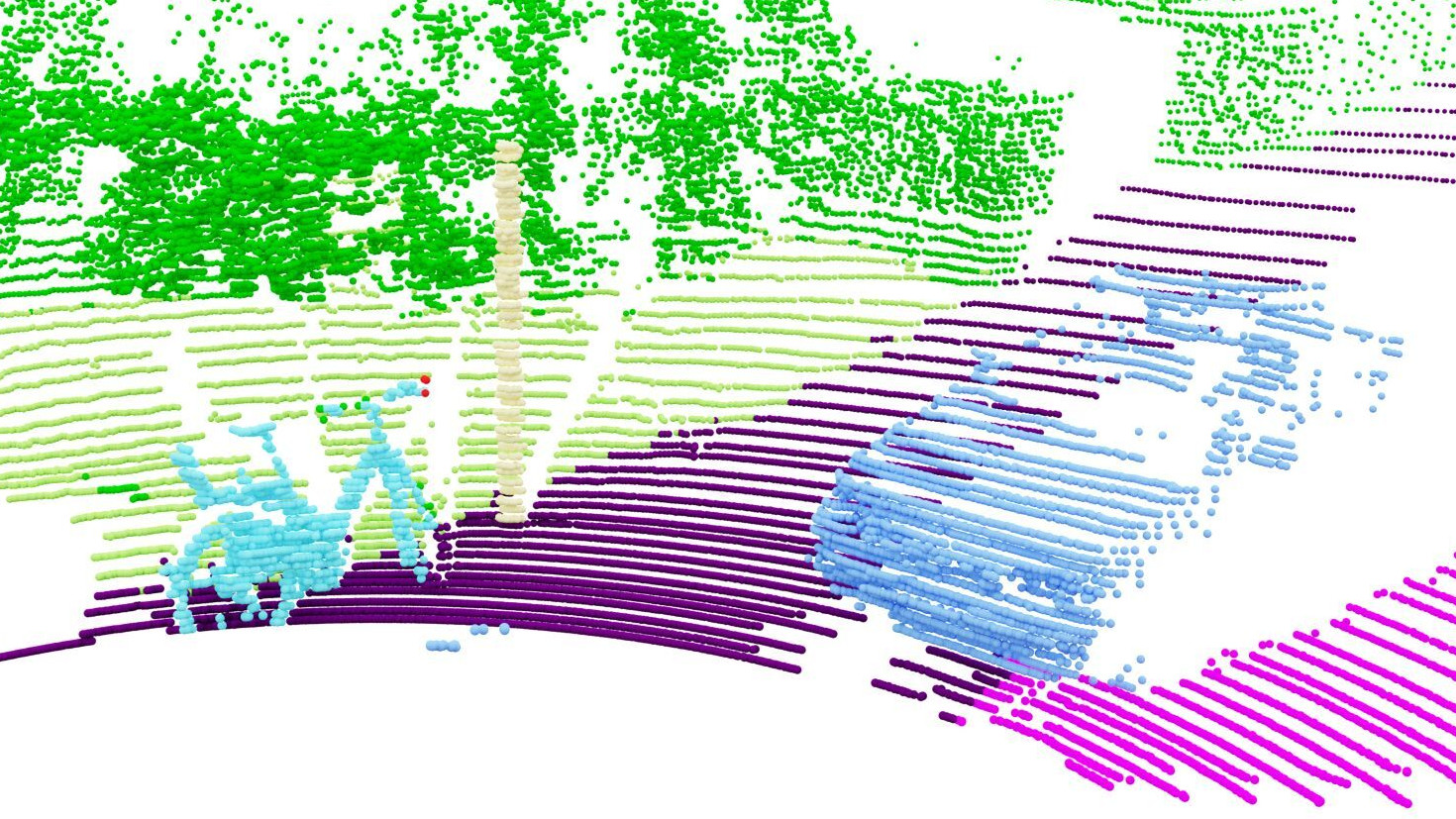}\\
    \vspace{-3px}
    \caption{\label{fig:s3dis} SemanticKITTI\,\cite{Behley19ICCV}}
\end{subfigure}
\hfill
\begin{subfigure}[b]{0.32\textwidth}
    \centering
            \includegraphics[width=\textwidth, trim={0 0 0 0}, clip]{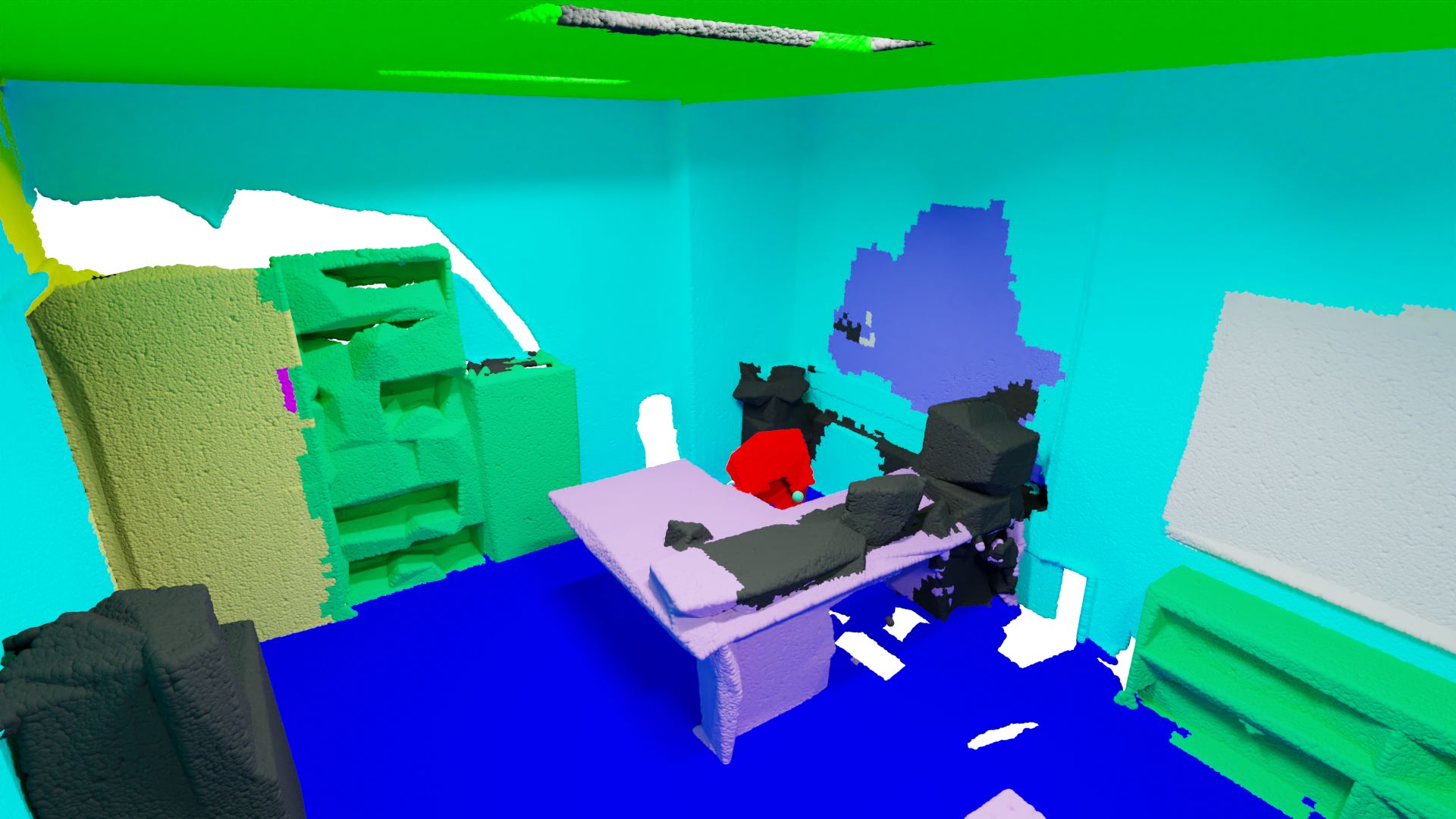}\\
    \includegraphics[width=\textwidth, trim={0 0 0 0}, clip]{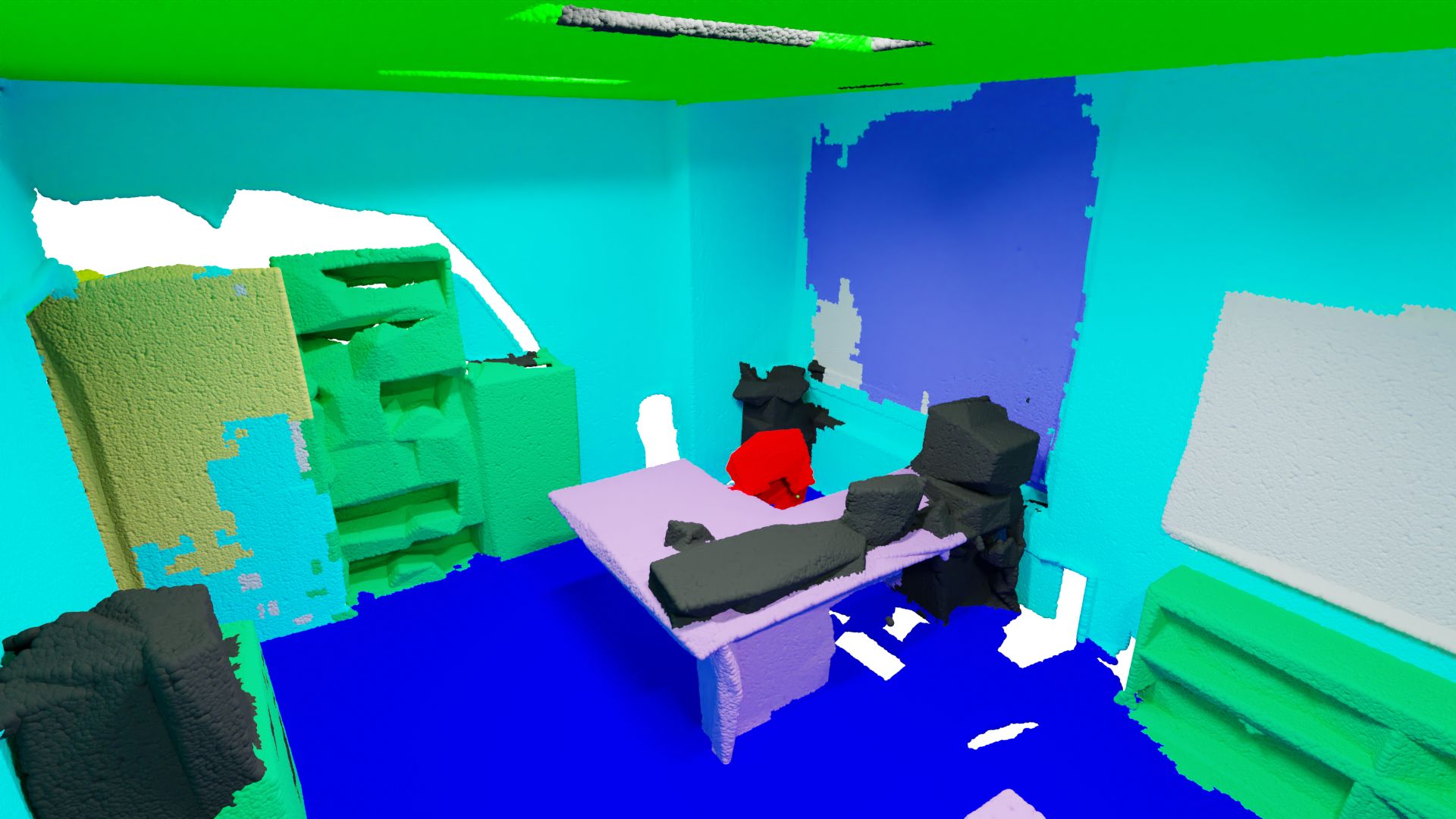}\\
    \vspace{-3px}
    \caption{\label{fig:scannet} S3DIS\,\cite{Armeni16CVPR}}
\end{subfigure}
\end{center}
\vspace{-18pt}
\caption{\textbf{Qualitative results.}
We show qualitative results of models trained with and without Mix3D on ScanNet \emph{(left)}, SemanticKITTI \emph{(center)} and S3DIS \emph{(right)}.
Compared to the original model, in \textbf{(a)} Mix3D helps to tell apart two stacked washing machines labeled as \textcolor{otherfurn}{\ColorMapCircle}~``other furniture" from \textcolor{refrigerator}{\ColorMapCircle}~``fridge".
In \textbf{(b)}, the MinkowskiNet trained without Mix3D wrongly classifies the \textcolor{bike}{\ColorMapCircle}~``bicycle" on the sidewalk next to the vegetation as \textcolor{other}{\ColorMapCircle}~``other-object".
For S3DIS, models trained with Mix3D particularly profit in the \textcolor{window}{\ColorMapCircle}~``window" class, as in \textbf{(c)}.
}
\vspace{-10pt}
\label{fig:qualitative_results}
\end{figure*}

\parag{How important is the scene overlap in Mix3D?}
The previous experiment established that mixing scenes is key to the improved performance of Mix3D.
In the next experiment, we take a closer look at the mixing itself, \ie, the combination of two scenes with increasing overlap.
The baseline is a MinkowskiNet trained without Mix3D, \ie, a single scene constitutes a training sample (\cf \reftab{scenes_nearby}, \circlenum{1}).
As Mix3D technically concatenates batch entries in pairs,
we conduct an experiment (\cf \reftab{scenes_nearby}, \circlenum{2}) to see the influence of the altered batching, \ie, the concatenation of two scenes that are placed very far away,
such that the receptive field of the model does not enable any information flow between the two scenes.
The scores are reported in \reftab{scenes_nearby}.
Perhaps not surprisingly, this setup shows no significant difference to the baseline, as contexts of the scenes do not mutually influence each other.
In the next setup (\cf \reftab{scenes_nearby}, \circlenum{3}), the scenes are placed nearby, such that information can flow via the receptive field, however without physical overlap of the scenes.
We observe an improvement of at least $+1.2$ mIoU compared to the baseline, as now each scene is influenced indirectly by the context of the neighboring scene via the receptive field of the model.
In the last experiment (\cf \reftab{scenes_nearby}, \circlenum{4}), we also include physical overlap of the scenes (this corresponds to Mix3D) and we observe the biggest improvement ($+2.4$ mIoU) compared to the baseline.
We conclude that the context overlap caused by mixing scenes is the decisive aspect which explains the performance of Mix3D.
In particular, objects of one scene are additionally exposed to the context of the other scene, and vice versa.

\begin{table}[h]
\begin{footnotesize}
\begin{overpic}[unit=1mm,width=8.4cm]{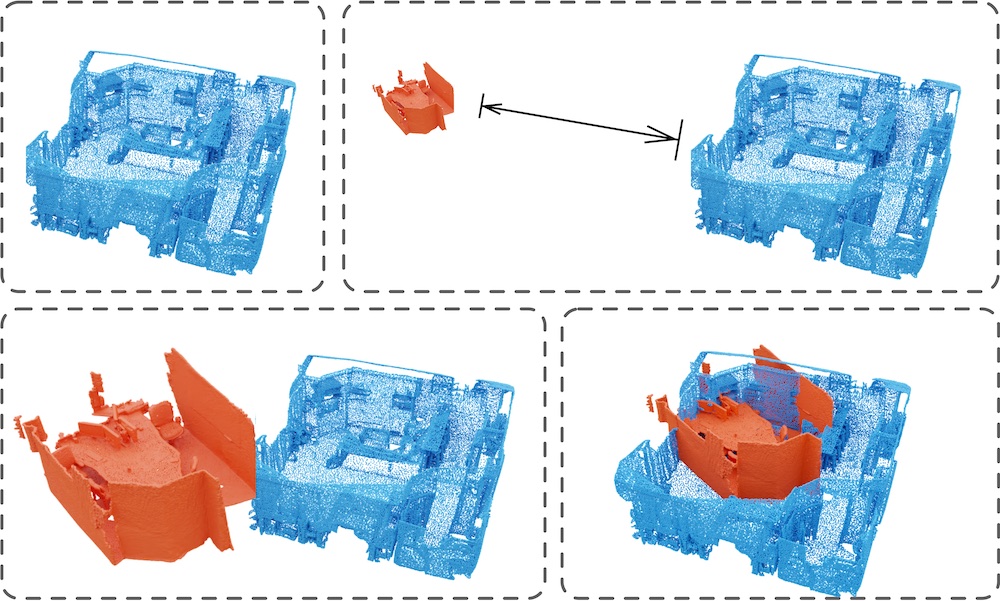}
\put(3, 56){\circlenum{\tiny 1} Single scene}
\put(37, 56){\circlenum{\tiny 2} No overlap (and far away)}
\put(3, 25.5){\circlenum{\tiny 3} No overlap (but nearby)}
\put(58, 25.5){\circlenum{\tiny 4} Overlap (\ie Mix3D)}
\end{overpic}
\end{footnotesize}
    \centering
    \begin{tabular}{l c c}
        \toprule
        Setup &\hspace{5px} & mIoU \\
        \midrule
        \circlenum{1} No mixing (single scene only) && 66.6 \\
        \circlenum{2} No overlap (outside receptive field) && 66.7 \\
        \circlenum{3} No overlap (inside receptive field) && \expNearby \\
        \circlenum{4} Overlap (\ie \namelong{}) && \textbf{69.0} \\
        \bottomrule
    \end{tabular}
\vspace{-5px}
\caption{\small \textbf{Influence of scene overlap.}
Overlap of scene-context is the decisive aspect of \namelong{}.
Placing scenes next to each other without overlaps or too far away for the model's receptive field to enable mutual information exchange between both scenes.
}
\vspace{0pt}
\label{tab:scenes_nearby}
\end{table}

\begin{figure}[t]
\centering
\input{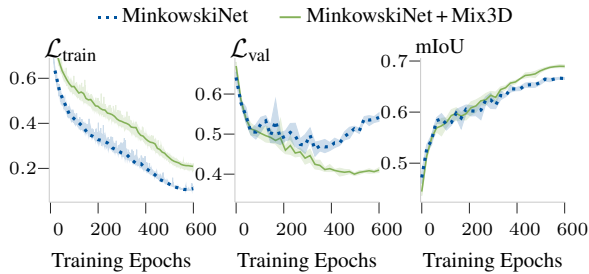}
\vspace{-10px}
\caption{\small \textbf{Learning curves with and without Mix3D.}
We show the mean and standard deviation over three training runs of a MinkowskiNet trained with and without Mix3D evaluated on the ScanNet validation set.
The MinkowskiNet is trained on $5$\,cm voxels including the standard data augmentations: rotation, translation, color-jitter and elastic distortion.
The training loss is notably lower without Mix3D \emph{(left)}.
The validation loss increases much later for models trained with Mix3D \emph{(center)} which indicates less overfitting, and results in overall improved performance \emph{(right)}.
}
\label{fig:overfitting}
\vspace{-19px}
\end{figure}

\parag{Does Mix3D improve generalization?}
Poor generalization can be caused by overfitting to the training set.
Overfitting and large generalization gaps are a considerable problem for 3D scene understanding tasks where training samples are few, and
in particular, when combined with high-capacity models\,\cite{Choy2019CVPR, Graham18CVPR, thomas19ICCV} that can overfit by memorizing the global context of the training scenes.
Learning curves are a useful tool to diagnose overfitting, specifically the generalization error on previously unseen data.
\reffig{overfitting} shows the training loss $\mathcal{L}_{\text{train}}$, validation loss $\mathcal{L}_{\text{val}}$ and the mean IoU per training epoch for models trained with and without Mix3D.
Mix3D reduces the generalization gap between $\mathcal{L}_{\text{train}}$ and $\mathcal{L}_{\text{val}}$, while $\mathcal{L}_{\text{val}}$ of the Mix3D models increases much later than for models trained without Mix3D.
This shows less overfitting to the training set and better generalization on the validation set in terms of semantic segmentation performance (mIoU).
Increasing the size of the training set is also a simple approach to reduce the generalization error \cite{Goodfellow16DLB}. 
Mix3D can be seen as a technique to create additional training examples by pairwise combination of existing samples.

\parag{Qualitative Results.}
\label{sec:qualitative_results}
We show qualitative comparisons of models trained with and without Mix3D. Specifically, in \reffig{teaser} we show out-of-context scenarios, and \reffig{qualitative_results} shows predictions on indoor and outdoor datasets.


\section{Discussion and Conclusion}
    \vspace{-3px}
In this work, we have introduced Mix3D, a simple yet effective data augmentation technique for large-scale 3D scene segmentation.
Models trained with Mix3D learn to better balance global context priors and local structure, as indicated by our analysis.
In experiments, we have shown that Mix3D significantly boosts the semantic segmentation performance of multiple state-of-the-art 3D deep learning models on both indoor and outdoor datasets.
Importantly, Mix3D is easy to implement and can directly be incorporated into existing training pipelines.
It remains an open research question how Mix3D can be applied to approaches combining 3D scans and corresponding 2D images.
Further, increased realism could improve the performance, since Mix3D generates unrealistic scenes.
Similarly, other tasks, such as object detection or instance segmentation, can be investigated.
In future research, we intend to investigate how to adapt the strong generalization abilities of Mix3D for weakly- and self-supervised learning, and we expect to see further developments along this line of work.
\vspace{-5px}
{\small
\paragraph{\small{Acknowledgments:}}
This work is supported by the ERC Consolidator Grant DeeViSe (ERC-2017-CoG-773161)
and compute resources from RWTH Aachen University (rwth0629, thes0775).
Contributions from the first author are part of his master thesis.
}

\newpage
\balance
{\small
\bibliographystyle{ieee}
\bibliography{abbrev,egbib}
}

\clearpage

\thispagestyle{empty}
 \twocolumn[{%
 \renewcommand\twocolumn[1][]{#1}
 \vspace{0.1cm}
 \begin{center}
 \textbf{\Large \papertitle \\
 \vspace{5px}
 \normalfont{Supplementary Material}}
 \end{center}
 \vspace{0.5cm}
 }]

\renewcommand{\thesection}{\Alph{section}}
\setcounter{section}{0}

\begin{abstract}
This supplementary material provides more details on the network architectures, including training and inference details (\refsec{architectures_training_details}).
In \refsec{analysis}, we provide additional analysis, experiments and results.
Finally, in \refsec{per_class_scores} we report per-class semantic segmentation scores on all three datasets (ScanNet, S3DIS, SemanticKITTI).
\end{abstract}

\section{Details on Architectures and Training}
\label{sec:architectures_training_details}
\paragraph{MinkowskiNet\,\cite{Choy2019CVPR}}
In \reftab{minkowskinet_architectures}, we provide the architectural details of the MinkowskiNet\,\cite{Choy2019CVPR} model used for the reported scores on the datasets ScanNet\,\cite{Dai17CVPR}, S3DIS\,\cite{Armeni16CVPR} and SemanticKITTI\,\cite{Behley19ICCV}.
We use the same model definitions as provided in the official code release of MinkowskiNet\textsuperscript{1}.
For ScanNet, we use $2$\,cm voxels for comparing to state-of-the-art methods and for the benchmark results (see \reftab{sota_comparison}-\ref{tab:scannet_benchmark}).
All other analysis experiments are performed on $5$\,cm voxels for faster training.

\paragraph{KPConv\,\cite{thomas19ICCV}}
We use the official TensorFlow code release\textsuperscript{2} for rigid and deformable KPConv on ScanNet and S3DIS. 
For rigid KPConv on SemanticKITTI, we utilize the PyTorch reimplementation\textsuperscript{3} by the original authors. 
Code for deformable KPConv on SemanticKITTI is not directly available.
In particular, we follow the same evaluation procedure as KPConv.
That is, we perform 100 evaluation runs for ScanNet and S3DIS (20 runs for SemanticKITTI) on augmented validation samples and conduct majority voting on the resulting predictions.
Note that we do not apply \namelong{} during test time, \ie, during inference the trained networks operate one single scenes.
\paragraph{Training details.}
As \namelong{} combines scenes, the total number of points in a single training sample consists of the sum over the number of all points of each mixed scene.
To ensure a fair comparison between MinkowskiNets, we appropriately reduce the batch size of \namelong{} for keeping the total number of points in a training iteration equal to the one of the baseline method.
For instance, we train the $2$\,cm baseline MinkowskiNet on a batch size of $6$.
When training with \namelong{}, we reduce the batch size to $3$ as each training sample consists of $2$ mixed scenes.
By doing so, we ensure that each model is trained on the same number of labeled points and therefore the results are comparable.
For KPConv, we resort to approximately keeping the total number of points in a train iteration equal.
Here, the original codebase operates on a variable batch size with a variable number of points in each sample.
We therefore occasionally need to reject samples for Mix3D in order to mix the specified number of spheres.
To ensure comparability to the baseline method, we follow the same training protocol proposed by Thomas \etal\,\cite{thomas19ICCV} and evaluate after the same number of epochs.

\footnotetext{\textsuperscript{1}\href{https://www.github.com/chrischoy/SpatioTemporalSegmentation-ScanNet}{www.github.com/chrischoy/SpatioTemporalSegmentation-ScanNet}}
\footnotetext{\textsuperscript{2}\href{https://www.github.com/HuguesTHOMAS/KPConv}{www.github.com/HuguesTHOMAS/KPConv}}
\footnotetext{\textsuperscript{3}\href{https://www.github.com/HuguesTHOMAS/KPConv-PyTorch}{www.github.com/HuguesTHOMAS/KPConv-PyTorch}}

\paragraph{Inference settings.}
For the ScanNet benchmark submission, we train an ensemble of four MinkowskiNet\,\cite{Choy2019CVPR} models with Mix3D.
In particular, the four models are Res16UNet34C and Res16UNet34A, each with the first convolution kernel size set to 3 and 5.
We train each model for $12 \cdot 10^{4}$ iterations on both the official training and validation splits of ScanNet\,\cite{Dai17CVPR}.
Similar to \cite{Graham18CVPR,Schult20CVPR,thomas19ICCV}, we perform test-time-augmentation (translation, rotation, scaling, color-jitter, random brightness and contrast) and average the predictions of 40 runs for each model.
To obtain the final predictions, we employ a graph-based over-segmentation scheme \cite{felzenszwalb2004efficient} (similar to OccuSeg\,\cite{Han20CVPR} and \cite{liu2021one}) with the goal to  
further smooth the predicted labels.

\begin{table}[b]
\resizebox{\columnwidth}{!}{%
    \begin{tabular}{r c c c c}
         \toprule
         Dataset & Voxel\,Size & Model & Input\,Features \\
         \midrule
         \multirow{2}{*}{ScanNet\,\cite{Dai17CVPR}} & $2$\,cm & Res16UNet34A & RGB \\
         & $5$\,cm & Res16UNet34C & RGB \\
         S3DIS\,\cite{Armeni16CVPR} & $5$\,cm & Res16UNet34C & RGB \\
         Sem.KITTI\,\cite{Behley19ICCV} & $15$\,cm & Res16UNet14A & Reflect., Dist.\\
         \bottomrule
    \end{tabular}
    }
    \caption{\small \textbf{MinkowskiNet\,\cite{Choy2019CVPR} architectures.}
    We provide architectural details about the MinkowskiNetworks used on ScanNet, S3DIS and SemanticKITTI.
    In particular, we use standard MinkowskiNets provided in the official repository and did not change the model definitions.
    Our ablation study provided in the main paper is performed on $5$\,cm voxels.
    \label{tab:minkowskinet_architectures}
    }
\end{table}

\section{Additional Analysis}
\begin{table}[t!]
    \begin{subfigure}[b]{.18\textwidth}
        \centering
        \includegraphics[width=\textwidth,
        trim={150 100 150 50},clip]{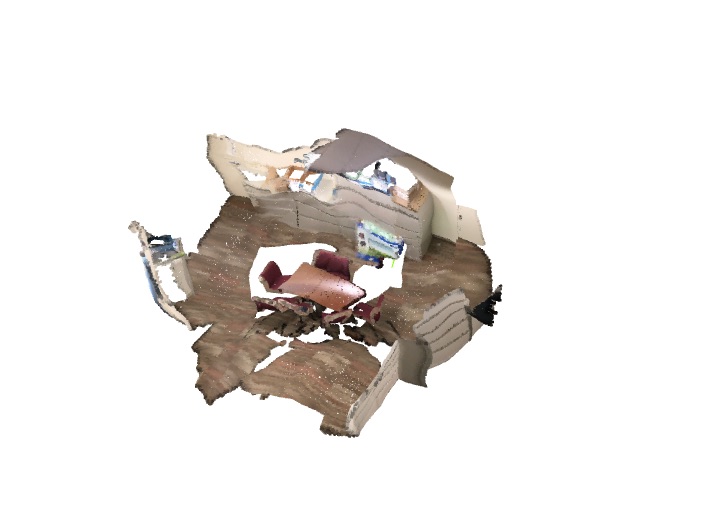}
        \caption{\small One Scene}
    \end{subfigure}
    \begin{subfigure}[b]{.18\textwidth}
        \centering
        \includegraphics[width=\textwidth,
        trim={100 160 100 50},clip]{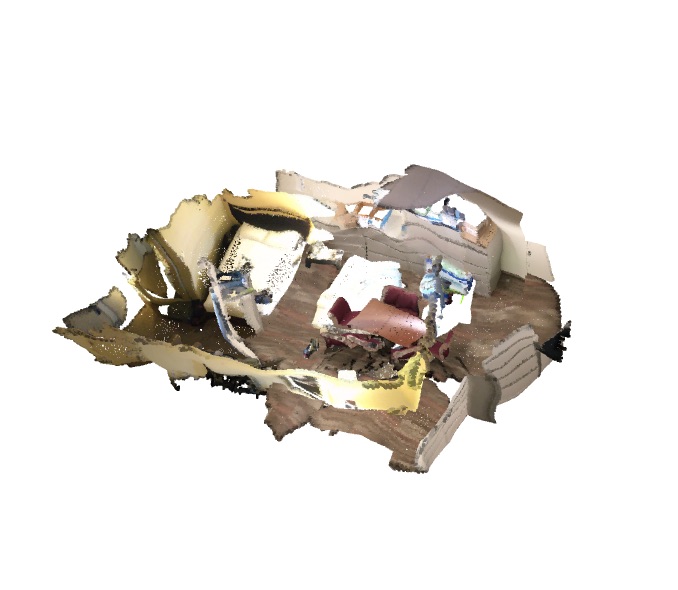}
        \caption{\small Two Scenes}
    \end{subfigure}
        \begin{subfigure}[b]{.18\textwidth}
        \centering
        \includegraphics[width=\textwidth,
        trim={150 100 150 50},clip]{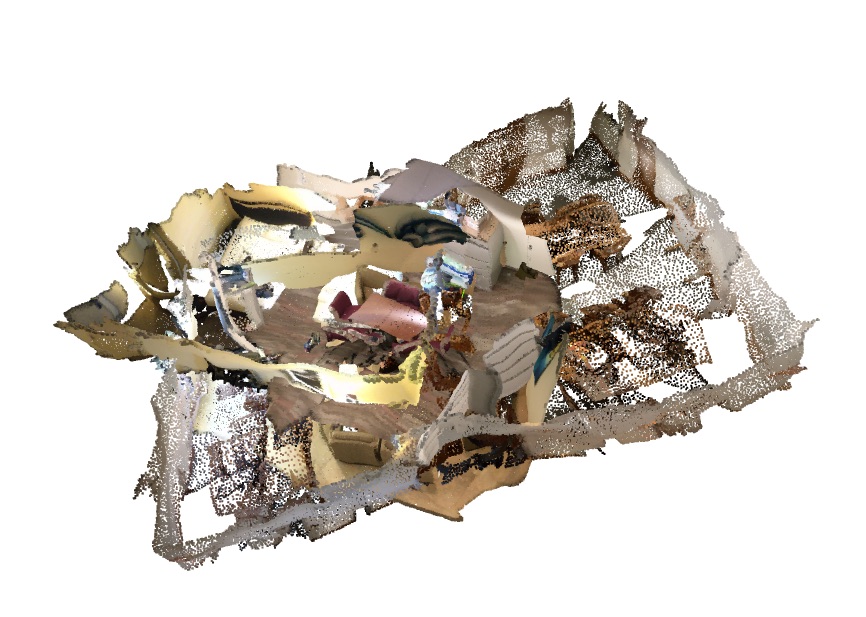}
        \caption{\small Four Scenes}
    \end{subfigure}
    \begin{subfigure}[b]{.18\textwidth}
        \centering
        \includegraphics[width=\textwidth,
        trim={100 160 100 50},clip]{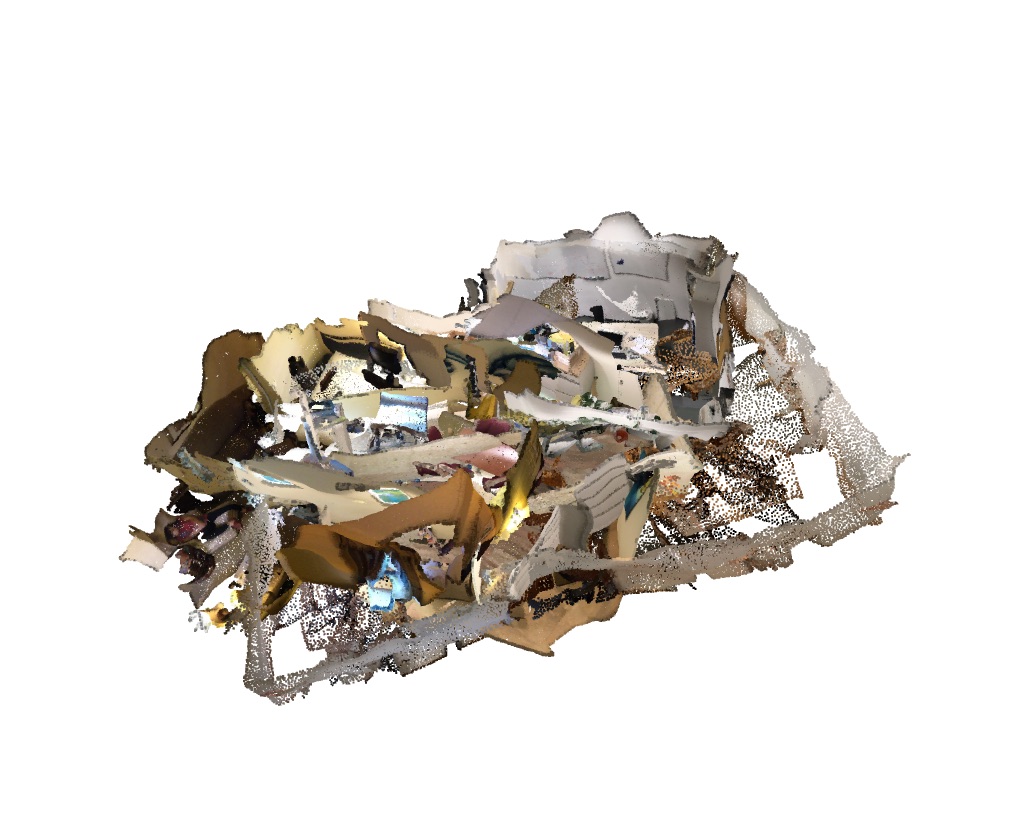}
        \caption{\small Eight Scenes}
    \end{subfigure}
    \centering
    \begin{tabular}{c | c c c c c c}
        \toprule
        \# Scenes & $1$ & $\mathbf{2}$ & $3$ & $4$ & $7$ & $8$ \\
        \midrule
        mIoU & $66.6$ & $\mathbf{69.0}$  &  $\expMergedThree$ & $\expMergedFour$ & $\expMergedSeven$ & $\expMergedEight$ \\
        \bottomrule
    \end{tabular}
   \vspace{-4pt}
    \caption{
    \label{tab:mixing_scenes}
    \small
    \textbf{Varying number of mixed point clouds.}
    While individual objects on two mixed point clouds \textbf{(b)} can still be easily distinguished,
    mixing eight point clouds \textbf{(d)} results in scenes where individual objects are hardly visually distinguishable.}
\end{table}
\label{sec:analysis_supp}

\paragraph{Influence of the number of mixed scenes.}
In this experiment, we vary the number of mixed scenes to evaluate if mixing more than two scenes further improves the segmentation performance.
\reftab{mixing_scenes} shows that the validation performance peaks at two mixed scenes and noticeably decreases with more scenes.
This is in line with other mixed sample data augmentation works\,\cite{Chen20ECCV,Verma19ICML, Zhang18ICLR}.

\paragraph{Influence of non-mixed spheres in a batch.}

On ScanNet, Thomas \etal\,\cite{thomas19ICCV} propose to train KPConv on spheres with $2$\,m radii cropped from the original point cloud.
As \namelong{} is implemented as concatenating subsequent pairs of batch entries, we will always obtain fully overlapping spheres in each novel training sample.
However, when training MinkowskiNets, we mix full \emph{scenes}.
This results in overlapping areas, but also in unmixed areas (\cf \reffig{mix3d}).
We therefore perform an ablation study with KPConv to measure the performance influence of adding non-mixed spheres to a batch, effectively imitating un-mixed regions.
In \reftab{vital2}, we observe that exposing the network also to non-mixed training samples improves the performance upon the initial \namelong{} implementation by up to $0.6$\,mIoU.

\paragraph{Consistent improvements wrt. dataset size.}
In \reftab{dataset_size}, we simulate a scarce data setting by training MinkowskiNet on randomly sampled subsets of the ScanNet train set and evaluate it on the full validation set.
We observe that \namelong{} achieves consistent absolute improvements ($abs. \Delta$) over all dataset sizes in the range of $[2.0, 2.8]$\,mIoU, while the relative improvement ($rel. \Delta$) becomes increasingly larger for smaller datasets. This shows the increased benefit of Mix3D when only few training samples are available.
\begin{table}[t]
    \centering
    \begin{tabular}{c c}
        \toprule
        \# non-mixed spheres & mIoU ($\pm$ $\sigma$) \\
        \midrule
        $0$ & $69.7$ $\pm$ $0.11$ \\
        $1$ & $70.1$ $\pm$ $0.35$\\
        $2$ & $70.3$ $\pm$ $0.06$ \\
        $3$ & $69.9$ $\pm$ $0.27$ \\
        $4$ & $70.3$ $\pm$ $0.11$ \\
        \midrule
        baseline & 69.3 $\pm$ 0.10 \\
        \bottomrule
    \end{tabular}
    \caption{
    \label{tab:vital2}
    \small \textbf{Influence of non-mixed spheres in a batch.}
    We observe that exposing the network also to non-mixed training samples has a significant impact on Deformable KPConv's performance on the ScanNet validation set\,\cite{Dai17CVPR}.
    Exposing the model only to mixed scenes achieves an improvement of $+0.4$\,mIoU, whereas having 2 or 4 non-mixed batch entries yields an improvement of $+1.0$\,mIoU. 
    Each batch contains about 10 spheres, depending on the number of points in each batch (\cite{thomas19ICCV} for details).
    }
\end{table}

\begin{table}[t]
    \centering
    \begin{tabular}{c c c c c}
    \toprule
    Dataset split & Baseline & $+$\name & $abs. \Delta$ & $rel. \Delta$ \\
    \midrule
    $5\,\%$   & $43.6$ & $45.6$ & $+2.0$ & $+4.6\,\%$ \\
    $25\,\%$  & $59.3$ & $62.1$ & $+2.8$ & $+4.7\,\%$ \\
    $50\,\%$  & $63.0$ & $65.7$ & $+2.7$ & $+4.3\,\%$ \\
    $75\,\%$  & $65.2$ & $67.3$ & $+2.1$ & $+3.2\,\%$ \\
    $100\,\%$ & $66.6$ & $69.0$ & $+2.4$ & $+3.6\,\%$ \\
    \bottomrule
    \end{tabular}
    \caption{\small \textbf{Consistent improvements wrt. dataset size.}
    We report consistent absolute performance gains in the range of $+[2.0, 2.8]$\,mIoU for all subsets of the ScanNet train set.}
    \label{tab:dataset_size}
\end{table}

\paragraph{Comparison to single-instance mixing.}
In~\reftab{single_instance_mixing}, we investigate an alternative formulation of Mix3D.
Instead of mixing complete scenes, we consider mixing single instances randomly sampled from an instance database.
This approach explicitly aligns with the key motivation of Mix3D -- reducing the context bias by placing instances in novel out-of-context environments.
Instance augmentation as proposed in 2D methods\,\cite{dwibedi17iccv, Ghiasi20arxiv} or 3D object detection\,\cite{yan18} has not yet been investigated for the task of 3D semantic segmentation.
As such, we suggest that mixing single instances constitutes an interesting avenue worth investigating.
To this end, we create a database consisting of instances from the training set.
Next, we place randomly sampled instances from this database in the scene.
In particular, we evaluate two strategies for instance placing: \circlenum{1} creating overlapping instances by placing randomly sampled instances exactly at the center position of arbitrary instances of the scene and \circlenum{2} randomly placing instances within the bounding scene volume.
We evaluate two sampling ratios.
For a sampling ratio of $+1.0$x, we sample exactly one other random instance from the instance database for each original instance in the scene.
For a sampling ratio of $+0.5$x, we sample half the number of original instances from the instance set.
We observe that single-instance mixing's performance is in the interval of three \namelong{} runs and is therefore comparable to Mix3D.
Perhaps unsurprisingly, single-instance mixing also improves significantly over the baseline,
as the key idea of instance mixing aligns with the motivation of Mix3D.
However, Mix3D is easier to implement into existing pipelines, and avoids the following fundamental disadvantages of single-instance mixing:
(1) In contrast to Mix3D, single-instance mixing requires a database of object instances available during training to augment the scenes.
(2) The construction of such a database requires instance level annotations which are not necessarily available in semantic segmentation datasets.
As an alternative, one could rely on synthetic CAD models (\eg from ShapeNet). Mix3D does not require additional annotations or other datasets.
(3) Single-Instance mixing requires a dedicated instance placing procedure introducing potentially sensitive hyper-parameters, while Mix3D simply concatenates overlapping scenes.
(4) Finally, by mixing the scene with instances, we effectively train on more data points, \ie, we introduce a compounding factor of exposing the model to instances of the training set more often as we do for \namelong{}. This makes a fair comparison harder.

\begin{table}[t]
\begin{overpic}[unit=1mm, scale=0.15]{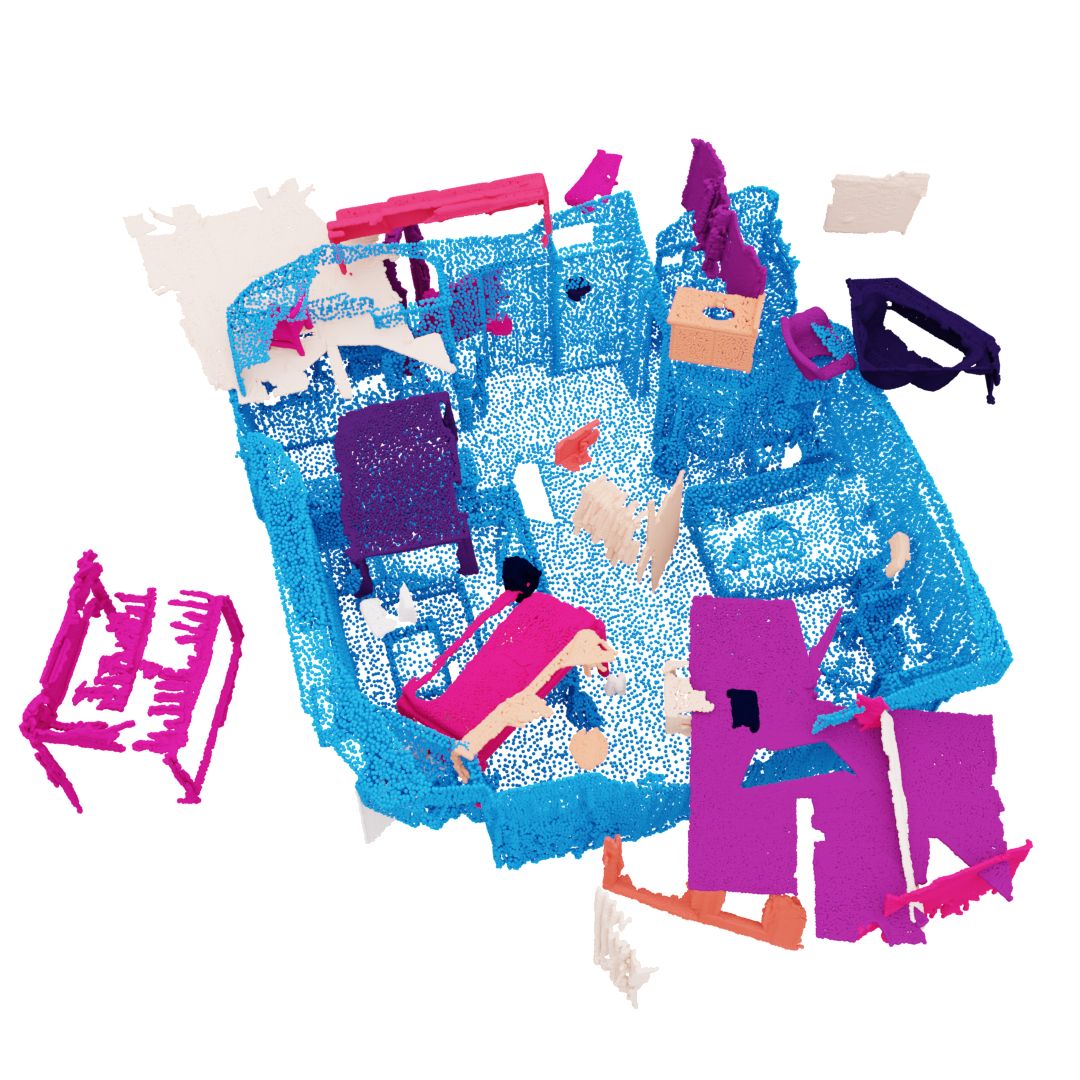}
\end{overpic}
    \centering
    \setlength{\tabcolsep}{4.5pt}{
    \begin{tabular}{c c c c}
        \toprule
        &placing method & sampling ratio & mIoU \\
        \midrule
        \multirow{2}{*}{\circlenum{1}} & \multirow{2}{*}{overlapping instances} & +1.0x  & \expPopulationHA \\
        && +0.5x & \expPopulationLA \\
        \midrule
        \multirow{2}{*}{\circlenum{2}} & \multirow{2}{*}{free placing} & +1.0x & \expPopulationHS \\
        && +0.5x & \textbf{\expPopulationLS} \\
        \midrule
            &\multicolumn{2}{c}{Baseline} & \expBaselineScannetMinkowski \\
            &\multicolumn{2}{c}{\namelong{}} & \textbf{\expMergedScannetMinkowski} \\
        \bottomrule
    \end{tabular}
    }
\vspace{-0px}
\caption{\small \textbf{Mixing with single instances.}
For every instance originally occurring in the scene we add 1x or 0.5x as many instances.
We place these newly sampled instances either at positions of already existing instances, \ie \emph{overlapping}, or freely in the volume of the scene, \ie, \emph{free space}, where we might create overlaps with existing instances. Above, we show an augmented scene with instances freely placed in the bounding volume with a sampling ratio of +1.0x.
We observe that single-instance mixing's performance is in the interval of three \namelong{} runs.
}
\vspace{-10pt}
\label{tab:single_instance_mixing}
\end{table}

\section{Per-Class Quantitative Evaluation}
\label{sec:per_class_scores}
\reftab{scannet_val_scores}-\ref{tab:semkitti_test_results} show per-class semantic segmentation results on all three datasets and different models trained with and without Mix3D.
Moreover, in \reftab{isolated_instances}, we extend \reftab{val_on_instances} and provide the per-class segmentation scores on isolated instances of the ScanNet validation set.
When no scene context is available, a model trained with Mix3D outperforms the same model trained without Mix3D on 18 out of 20 semantic classes, indicating that Mix3D enables models to extract more meaningful information from local geometry and structures alone.

\begin{table*}[h!]
\resizebox{\textwidth}{!}{%
\setlength{\tabcolsep}{1.8pt}
\begin{tabular}{rcc|cccccccccccccccccccc}
        \toprule
        Method & \namelong{} & mIoU & wall & floor & cabinet & bed & chair & sofa & table & door & wndw & books. & pic & counter & desk & curtain & fridge & shower & toilet & sink & bathtub & otherf.\\
        \midrule
        MinkNet\,\cite{Choy2019CVPR} & \xmark & 66.7 & 81.0 & 93.8 & 61.0  & \textbf{79.7} & 88.6 & 79.6 & \textbf{71.9} & 55.6 & 57.54 & 75.4 & 25.4 & 55.0 & 61.3 & 61.3 & 44.2 & 61.6 & 85.9 & 61.8 & 79.3 & 53.8 \\
        (5cm) & \cmark & \textbf{69.1} & \textbf{82.1} & \textbf{93.9} & \textbf{63.9} & 79.2 & \textbf{89.2} & \textbf{81.1} & 70.8 & \textbf{59.3} & \textbf{59.8} & \textbf{78.1} & \textbf{25.6} & \textbf{62.2} & \textbf{62.1} & \textbf{67.0} & \textbf{54.1} & \textbf{64.0} & \textbf{89.8} & \textbf{64.1} & \textbf{81.1} & \textbf{54.3}\\
        \midrule
        MinkNet\,\cite{Choy2019CVPR} & \xmark & 72.4 & 85.6 & 96.5 & 64.7 & \textbf{82.1} & 91.0 & 84.4  & 74.5 & \textbf{65.0} & 62.8 & 79.5 & 32.4 & 64.4 & 63.7  & \textbf{75.5} & 51.6  &  \textbf{69.0}  & 93.0  & 67.6 & 87.6  & \textbf{56.3} \\
        (2cm) & \cmark & \textbf{73.6}  & \textbf{86.0}  & \textbf{96.6} & \textbf{66.3} & \textbf{82.1} & \textbf{91.9} & \textbf{86.1} & \textbf{75.7}  & 64.7 & \textbf{64.7} & \textbf{79.6} & \textbf{36.3} & \textbf{67.7} & \textbf{67.0} & 74.7 & \textbf{59.3} & 68.8 & \textbf{93.9} & \textbf{68.6} & \textbf{87.9} & 55.1 \\
        \midrule
        KPConv\,\cite{thomas19ICCV} & \xmark & 68.8 & 82.2 & 94.4 & 63.4 & 77.9 & 88.3 & 76.6 & \textbf{72.8} & 60.0 & 58.4 & 80.1 & 28.6 & 59.3 & 62.7 & 69.9 & 53.1 & \textbf{51.5} & 90.6 & 63.4 & \textbf{86.4} & 55.3 \\
        (Rigid) & \cmark & \textbf{69.5} & \textbf{82.5} & \textbf{94.6} & \textbf{65.3} & \textbf{79.8} & \textbf{89.7} & \textbf{79.0} & 72.2 & \textbf{61.0} & \textbf{58.7} & \textbf{80.1} & \textbf{30.6} & \textbf{60.0} & \textbf{62.8} & \textbf{70.7} & \textbf{55.3} & 50.2 & \textbf{91.5} & \textbf{65.0} & 85.3 & \textbf{55.6} \\
        \midrule
        KPConv\,\cite{thomas19ICCV} & \xmark & 69.3 & 82.4 & \textbf{94.4} & 64.5 & 79.2 & 88.5 & 77.2 & \textbf{73.0} & 60.5 & \textbf{59.1} & \textbf{79.8} & 28.4 & 59.9 & \textbf{63.7} & 71.6 & 53.1 & 54.1 & 91.5 & 63.3 & \textbf{86.0} & \textbf{56.4} \\
        (Deformable) & \cmark & \textbf{70.3} & \textbf{82.6} & \textbf{94.4} & \textbf{65.8} & \textbf{79.5} & \textbf{90.2} & \textbf{79.9} & 72.5 & \textbf{61.7} & 58.2 & 79.6 & \textbf{31.5} & \textbf{62.1} & 63.2 & \textbf{72.0} & \textbf{58.1} & \textbf{56.8} & \textbf{91.8} & \textbf{64.6} & 85.6 & 55.4 \\
        \bottomrule
\end{tabular}
}
\vspace{-6pt}
\caption{
\small
\label{tab:scannet_val_scores}
\textbf{ScanNet\,\cite{Dai17CVPR} validation set.}
Per-class semantic segmentation IoU scores. 
We report the mean performance on the validation set over 3 trained models.
}
\end{table*}

\begin{table*}[h!]
\resizebox{\textwidth}{!}{%
\setlength\tabcolsep{3pt}
\begin{tabular}{rcc|ccccccccccccc}
\toprule
Method	 & \namelong{} & mIoU & ceiling	 & floor	 & wall	 & beam	 & column	 & window & door	 & chair	 & table	 & bookshelf	 & sofa	 & board & clutter \\
\midrule
MinkNet\,\cite{Choy2019CVPR} & \xmark & 64.8 & 92.6 & 96.4 & 82.0 & 0.2 & \textbf{28.1} & 53.3 & 64.1 & 87.7 & 78.5 & \textbf{73.8} & \textbf{65.5} & 63.1 & 56.4 \\
(5cm)& \cmark & \textbf{65.5} & \textbf{93.5} & \textbf{97.5} & \textbf{82.8} & 0.0 & 24.3 & \textbf{57.5} & \textbf{64.8} & \textbf{88.6} & \textbf{80.7} & 72.9 & 62.6 & \textbf{68.3} & \textbf{57.5} \\
\midrule
KPConv\,\cite{thomas19ICCV} & \xmark & 65.0 & 93.6 & 98.3 & 81.9 & 0.0 & \textbf{19.4} & 54.8 & 64.7 & 91.1 & 81.4 & 74.4 & 64.2 & 63.0 & 58.0 \\
(Rigid)& \cmark & \textbf{66.5} & \textbf{93.9} & \textbf{98.4} & \textbf{82.9} & 0.0 & \textbf{19.4} & \textbf{59.4} & \textbf{69.0} & \textbf{91.3} & \textbf{81.6} & \textbf{75.6} & \textbf{65.7} & \textbf{68.5} & \textbf{59.0} \\
\midrule
KPConv\,\cite{thomas19ICCV} & \xmark & 65.8 & 93.8 & 98.4 & 82.2 & 0.0 & 20.7 & 55.8 & \textbf{68.9} & 91.3 & 81.9 & 75.3 & 64.9 & 63.5 & 58.8 \\
(Deformable)& \cmark & \textbf{67.2} & \textbf{93.9} & \textbf{98.4} & \textbf{83.5} & 0.0 & \textbf{22.3} & \textbf{60.8} & 66.3 & \textbf{91.7} & \textbf{82.3} & \textbf{76.2} & \textbf{66.7} & \textbf{70.7} & \textbf{60.5} \\
\bottomrule
\end{tabular}
}
\vspace{-6pt}
\caption{
\small
\label{tab:Table_S3DIS_area5} 
\textbf{S3DIS\,\cite{Armeni16CVPR} Area $\mathbf{5}$.}
Per-class semantic segmentation IoU scores. 
We report the mean over 3 trained models.
}
\end{table*}

\begin{table*}[h!]
\resizebox{\textwidth}{!}{%
\setlength{\tabcolsep}{1.2pt}
\begin{tabular}{rcc|cccccccccccccccccccc}
        \toprule
        Method & \namelong{} & mIoU & car & bike & mbike & truck & vehicle & person & cyclist & mcyclist & road & parking & sidewalk & other-gr & building & fence & veget & trunk & terrain & pole & sign \\
        \midrule
        \multirow{2}{*}{MinkNet\,\cite{Choy2019CVPR}} & \xmark & 53.2 & 94.0 & 26.4 & 24.5 & 27.5 & 18.4 & 40.5 & 46.7 & 13.5 & 88.4 & 57.1 & 71.4 & 22.6 & 90.4 & 62.5 & \textbf{83.5} & 65.3 & \textbf{65.8} & 54.0 & 59.1 \\
        & \cmark & \textbf{58.1} & \textbf{95.1} & \textbf{30.6} & \textbf{38.3} & \textbf{43.5} & \textbf{35.3} & \textbf{50.8} & \textbf{51.8} & \textbf{21.9} & \textbf{89.4} & \textbf{59.5} & \textbf{73.0} & \textbf{23.2} & \textbf{90.9} & \textbf{65.2} & 83.3 & \textbf{68.0} & 65.4 & \textbf{57.6} & \textbf{61.7} \\
        \midrule
        \multirow{2}{*}{KPConv\,\cite{thomas19ICCV}} & \xmark & 63.1 & 95.9 & 45.2 & 50.9 & 44.4 & 45.4 & 62.9 & 64.7 & 36.6 & \textbf{89.4} & 65.1 & 73.5 & \textbf{31.3} & \textbf{91.7} & \textbf{67.5} & \textbf{84.4} & 69.4 & \textbf{68.4} & 57.9 & 54.4 \\
        & \cmark & \textbf{63.6} & \textbf{96.4} & \textbf{49.1} & \textbf{59.7} & \textbf{40.5} & \textbf{52.1} & \textbf{69.6} & \textbf{69.3} & \textbf{18.2} & \textbf{89.4} & \textbf{66.2} & \textbf{74.3} & 28.0 & 91.5 & 66.2 & \textbf{84.4} & \textbf{71.0} & 67.9 & \textbf{58.7} & \textbf{54.9} \\
        \bottomrule
\end{tabular}
}
\vspace{-6pt}
\caption{
\small
\label{tab:semkitti_test_results}
\textbf{SemanticKITTI\,\cite{Behley19ICCV} test set.}
Per-class semantic segmentation IoU scores.
}
\end{table*}

\begin{table*}[h!]
\resizebox{\textwidth}{!}{%
\setlength{\tabcolsep}{1.2pt}
\begin{tabular}{cc|cccccccccccccccccccc}
        \toprule
        \namelong{} & mIoU & wall & floor & cabinet & bed & chair & sofa & table & door & wndw & books. & pic & counter & desk & curtain & fridge & shower & toilet & sink & bathtub & otherf.\\
        \midrule
        \multirow{2}{*}{\xmark} & 24.58 & 57.95 & 85.76 & 6.29 & 34.64 & 65.90 & 66.28 & 13.91 & 3.26 & 20.61 & 52.65 & 0.78 & 0.26 & 8.63 & 19.72 & \textbf{3.89} & \textbf{0.11} & 10.65 & 7.51 & 16.28 & 16.47 \\
        & \small $\pm$0.89 & \small$\pm$0.97 & \small$\pm$0.43 & \small$\pm$0.56 & \small$\pm$4.46 & \small $\pm$5.36 &  \small$\pm$3.70 & \small$\pm$6.40 & \small$\pm$2.11 & \small$\pm$3.46 & \small$\pm$2.96 & \small$\pm$0.74 & \small$\pm$0.34 & \small$\pm$3.92 & \small$\pm$1.23 & \small$\pm$5.50 & \small$\pm$0.10 & \small$\pm$6.59 & \small$\pm$5.66 & \small$\pm$4.30 & \small$\pm$2.38 \\
        \midrule
        \multirow{2}{*}{\cmark} & \textbf{36.00} & \textbf{65.09} & \textbf{87.40} & \textbf{10.73} & \textbf{43.13} & \textbf{80.91} & \textbf{79.49} & \textbf{33.89} & \textbf{6.49} & \textbf{27.94} & \textbf{59.35} & \textbf{6.35} & \textbf{0.51} & \textbf{19.28} & \textbf{40.71} & 3.08 & 0.00 & \textbf{52.91} & \textbf{17.55} & \textbf{60.42} & \textbf{24.76} \\
        & \small$\pm$0.58 & \small$\pm$0.33 &  \small$\pm$1.91 & \small$\pm$0.81 & \small$\pm$1.74 & \small$\pm$0.92 & \small$\pm$1.42 & \small$\pm$4.98 & \small$\pm$0.50 & \small$\pm$4.46 & \small$\pm$0.98 & \small$\pm$4.36 & \small$\pm$0.49 & \small$\pm$6.51 & \small$\pm$1.27 & \small$\pm$1.30 & \small$\pm$0.00 & \small$\pm$8.65 & \small$\pm$8.90 & \small$\pm$6.27 & \small$\pm$3.23 \\
        \bottomrule
\end{tabular}
}
\vspace{-6pt}
\caption{
\small
\label{tab:isolated_instances}
\textbf{Isolated instances.}
Per-class semantic segmentation IoU scores on the ScanNet validation set\,\cite{Dai17CVPR}.
We extend Table 2 of the main paper by providing per-class semantic segmentation scores.
We report performance improvements on 18 out of 20 semantic classes.
The study is performed with MinkowskiNet on $5$\,cm voxels and we train 3 models to obtain the means and standard deviations.
}
\end{table*}

\end{document}